\documentclass[sigconf,screen]{acmart}

\usepackage{overpic}
\usepackage{overpic} 
\usepackage{color}
\usepackage{mathtools} 
\usepackage{currfile}
\usepackage{multirow}

\definecolor{turquoise}{cmyk}{0.65,0,0.1,0.3}
\definecolor{purple}{rgb}{0.65,0,0.65}
\definecolor{dark_green}{rgb}{0, 0.5, 0}
\definecolor{orange}{rgb}{0.8, 0.6, 0.2}
\definecolor{red}{rgb}{0.8, 0.2, 0.2}
\definecolor{darkred}{rgb}{0.6, 0.1, 0.05}
\definecolor{blueish}{rgb}{0.0, 0.3, .6}
\definecolor{light_gray}{rgb}{0.7, 0.7, .7}
\definecolor{pink}{rgb}{1, 0, 1}
\definecolor{greyblue}{rgb}{0.25, 0.25, 1}

\renewcommand{\paragraph}[1]{\vspace{.5em}\noindent\textbf{#1}.}

\usepackage{lipsum}

\usepackage{blindtext}

\newcommand{\kostas}[1]{{\color{Bittersweet} {[\bf Kosta: #1]}}}
\newcommand{\andrew}[1]{{\color{BlueViolet} {[Andrew: #1]}}}
\newcommand{\vitto}[1]{{\color{OrangeRed} {[Vitto: #1]}}}
\newcommand{\JB}[1]{{\color{OliveGreen} {[Jon: #1]}}}
\newcommand{\tom}[1]{{\color{RoyalPurple} {[Tom: #1]}}}
\newcommand{\pratul}[1]{{\color{Emerald} {[Pratul: #1]}}}
\newcommand{\at}[1]{{\color{blueish}#1}}
\newcommand{\AT}[1]{{\color{blueish}{\bf [Andrea: #1]}}}
\newcommand{\At}[1]{\marginpar{\tiny{\textcolor{blueish}{#1}}}}
\newcommand{\al}[1]{\textbf{\color{orange}[AL: #1]}}
\renewcommand{\kostas}[1]{}
\renewcommand{\andrew}[1]{}
\renewcommand{\vitto}[1]{}
\renewcommand{\JB}[1]{}
\renewcommand{\tom}[1]{}
\renewcommand{\pratul}[1]{}
\renewcommand{\at}[1]{}
\renewcommand{\AT}[1]{}
\renewcommand{\At}[1]{}
\renewcommand{\al}[1]{}

\def\eg{\textit{e.g.,~}}               
\def\ie{\textit{i.e.,~}}      
\def\etc{\textit{etc}}                 

\newlength\paramargin
\newlength\figmargin

\newlength\secmargin
\newlength\figcapmargin
\newlength\tabcapmargin

\newcommand{\topic}[1]
{
\vspace{1.5mm}\noindent\textbf{#1}
}

\long\def\ignorethis#1{}

\newbox\jsavebox%

\makeatletter
\newcommand{\providelength}[1]{%
  \@ifundefined{\expandafter\@gobble\string#1}
   {
    \typeout{\string\providelength: making new length \string#1}%
    \newlength{#1}%
   }
   {
    \sdaau@checkforlength{#1}%
   }%
}

\newcommand{\tabref}[1]{Tab.~\ref{#1}}
\newcommand{\equref}[1]{Eq.~\eqref{#1}}
\newcommand{\figref}[1]{Fig.~\ref{#1}}
\newcommand{\secref}[1]{Sec.~\ref{#1}}

\def\suppurl{https://cwchenwang.github.io/outdoor-nerf-depth/data/supp.pdf}

\usepackage{pifont}
\usepackage{caption}

\AtBeginDocument{%
  }

\copyrightyear{2023}
\acmYear{2023}
\setcopyright{acmlicensed}\acmConference[MM '23]{Proceedings of the 31st
ACM International Conference on Multimedia}{October 29-November 3,
2023}{Ottawa, ON, Canada}
\acmBooktitle{Proceedings of the 31st ACM International Conference on
Multimedia (MM '23), October 29-November 3, 2023, Ottawa, ON, Canada}
\acmPrice{15.00}
\acmDOI{10.1145/3581783.3612306}
\acmISBN{979-8-4007-0108-5/23/10}

\acmSubmissionID{2698}

\begin{document}

\title{Digging into Depth Priors for Outdoor Neural Radiance Fields}

\author{Chen Wang}
\orcid{0000-0002-9315-3780}
\email{cw.chenwang@outlook.com}
\affiliation{%
  \institution{Tsinghua University, Baidu Research}
  \state{Beijing}
  \country{China}
}
\author{Jiadai Sun}
\orcid{0000-0001-9593-0873}
\affiliation{%
  \institution{Northwestern Polytechnical University, Baidu Research}
  \state{Beijing}
  \country{China}
}

\author{Lina Liu}
\orcid{0000-0001-6483-1557}
\affiliation{%
  \institution{Zhejiang University, Baidu Research}
  \state{Beijing}
  \country{China}
}

\author{Chenming Wu}
\authornote{Corresponding authors (\{wuchenming, shenzhelun\}@baidu.com).}
\orcid{0000-0001-8012-1547}
\affiliation{%
  \institution{RAL, Baidu Research}
  \state{Beijing}
  \country{China}
}

\author{Zhelun Shen}
\orcid{0009-0006-6540-856X}
\authornotemark[1]
\affiliation{%
  \institution{RAL, Baidu Research}
  \state{Beijing}
  \country{China}
}

\author{Dayan Wu}
\orcid{0000-0002-8604-7226}
\affiliation{%
  \institution{Institute of Information Engineering, Chinese Academy of Sciences}
  \state{Beijing}
  \country{China}
}

\author{Yuchao Dai}
\orcid{0000-0002-4432-7406}
\affiliation{%
  \institution{Northwestern Polytechnical University}
  \state{Xi'an}
  \country{China}
}

\author{Liangjun Zhang}
\orcid{0000-0001-5737-2540}
\affiliation{%
  \institution{RAL, Baidu Research}
  \state{Sunnyvale, CA}
  \country{USA}
}

\renewcommand{\shortauthors}{Wang et al.}

\begin{abstract}
Neural Radiance Fields (NeRFs) have demonstrated impressive performance in vision and graphics tasks, such as novel view synthesis and immersive reality. However, the shape-radiance ambiguity of radiance fields remains a challenge, especially in the sparse viewpoints setting. Recent work resorts to integrating depth priors into outdoor NeRF training to alleviate the issue. However, the criteria for selecting depth priors and the relative merits of different priors have not been thoroughly investigated. Moreover, the relative merits of selecting different approaches to use the depth priors is also an unexplored problem. In this paper, we provide a comprehensive study and evaluation of employing depth priors to outdoor neural radiance fields, covering common depth sensing technologies and most application ways. Specifically, we conduct extensive experiments with two representative NeRF methods equipped with four commonly-used depth priors and different depth usages on two widely used outdoor datasets. Our experimental results reveal several interesting findings that can potentially benefit practitioners and researchers in training their NeRF models with depth priors. Project page: \href{https://cwchenwang.github.io/outdoor-nerf-depth}{https://cwchenwang.github.io/outdoor-nerf-depth}
\end{abstract}



\maketitle

\section{Introduction}\label{sec:intro}

\setlength{\textfloatsep}{1pt}
\begin{figure}[htbp]
    \centering
    \includegraphics[width=1.0\linewidth]{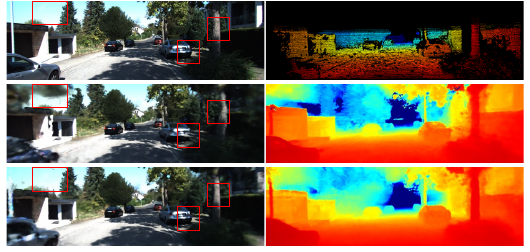}
    \caption{Image and depth map visualization of ground truth (top), trained with pure RGB (middle) and trained with monocular depth estimation (bottom), from a testing viewpoint. Even with a monocular depth (the quality is the worst compared to other depth priors), the view synthesis can be significantly improved for NeRF in terms of fewer floaters, and better preservation of object shapes (cars or trees) compared with using only RGB frames.}
    \label{fig:teaser}
\end{figure}

Novel view synthesis, \ie synthesizing photorealistic images at arbitrary viewpoints, is a long-standing task in computer vision and multimedia. Recently, neural radiance fields (NeRF)~\cite{mildenhall2020nerf} and its variants have been a new selection for novel view synthesis and achieved impressive performance. Specifically, NeRF represents a 3D scene by a continuous function, which takes a pair of 3D position and 2D viewing direction as input to predict RGB color and volume density. This enables us to render an image using standard volume rendering equation~\cite{kajiya1984ray}. The photorealistic renderings from NeRF motivate a large amount of recent work in multimedia area~\cite{zhang2022vmrf, yu2022pvserf, wu2022dof, xing2022mvsplenoctree, wang2022nerf, wang2023benchmarking}.


Despite the impressive development of NeRF, defining a reasonable underlying geometry from the radiance field is still an unresolved issue. To tackle this problem, several works use distance fields~\cite{wang2021neus, long2022neuraludf} to explicitly define the geometry for NeRF frameworks. However, these methods require a large number of input images from different perspectives (normally 360-degree capturing is required) to successfully reconstruct an object. In outdoor scenes, there exist many foreground and background objects, and fully capturing all of them is a difficult task. 
Using geometry priors, in particular depth priors, to facilitate outdoor NeRF training is necessary and has been proven effective in previous work~\cite{rematas2022urban,wang2023neural,meuleman2023progressively,anonymous2023snerf}.
Specifically, raw LiDAR depth, depth completion, and depth estimation are commonly used depth priors. Since they come at different costs and vary in accuracy, it is important to further investigate the criteria for selecting depth priors and the relative merits of them.

In this paper, we provide a comprehensive study and evaluation of fusing a different modal input, \ie depth priors, to outdoor neural radiance fields, covering common depth sensing technologies and most application ways in the all-time study of neural radiance fields. Specifically, depth sensing in outdoor scenes can be classified into two categories: (1) Active depth sensing: methods that employ optical devices to acquire depth. One major limitation is that the optical devices, \eg Light Detection and Ranging (LiDAR) are cumbersome and provide only sparse measurements. Depth completion is thus proposed to recover a dense depth map from a sparse one. (2) Passive depth sensing: methods directly infer depth from images, which is a much cheaper choice but sacrifices accuracy. Monocular and binocular depth estimation are two common methods. Experimentally, we select two representative NeRF methods and augment them with different depth supervision and different loss functions on two popular outdoor autonomous driving datasets: KITTI~\cite{geiger2012KITTI} and Argoverse~\cite{wilson2023argoverse}. As a result, we conclude the experimental results and have interesting findings as follows:

\begin{itemize}
    \item \textbf{Density}: Even a very sparse depth supervision can significantly boost the view synthesis quality, and generally the denser the better.
    \item \textbf{Quality}: (1)	Monocular depth is enough for sparse view inputs, which can even achieve comparable results with the ground truth depth supervision; (2) depth supervision is an option for dense view, \ie it is necessary if the corresponding application needs the employed NeRF to have a better geometry, such as 3D reconstruction.
    \item \textbf{Supervision}: Complex depth filtering and loss function is unnecessary in outdoor NeRF and directly cropping the sky area with MSE supervision is enough.
\end{itemize}
To the best of our knowledge, our work is the first quantitative and qualitative comparison of employing depth priors to outdoor neural radiance fields and we believe our findings would be helpful for practitioners and researchers to have a bigger picture of how to effectively incorporate depth priors in training outdoor NeRFs.

\section{Related Work}
\label{sec:related-work}
\topic{Neural Radiance Fields.}
Neural radiance fields (NeRFs)~\cite{mildenhall2021nerf} demonstrate superior effectiveness in novel view synthesis by predicting the per-point color and radiance of a 3D scene with a multi-layer perceptron (MLP). However, vanilla NeRF requires hours of optimization and assumes static scenes along with dense viewpoints. The following works have extended it in different aspects, \eg modeling dynamic and deformable scenes~\cite{park2021nerfies}, super-resolution~\cite{wang2022nerf}, sparse or imperfect poses~\cite{lin2021barf}, generalization to target scenes~\cite{yu2021pixelnerf} and fast optimization~\cite{yu2021plenoxels}. To enable using NeRF in unbounded outdoor scenes, NeRF++~\cite{zhang2020nerf++} introduces inverted sphere parametrization to handle unbounded scenes. MipNeRF-360~\cite{barron2022mip} re-parametrizes their scene coordinates with inverse-depth spacing, achieving evenly-spaced ray intervals in unbounded regions. In terms of large outdoor scenes, Block-NeRF~\cite{tancik2022block} decomposes the scene into multiple blocks and trains a NeRF individually. 
Recent work~\cite{wu23iros,yang2023unisim} validates the applicability of outdoor NeRF to autonomous driving simulation.

\topic{Depth Supervision for Neural Representations.}
Although vanilla NeRF only needs RGB images for training, when input viewpoints are sparse, the optimization can easily fall into local minima due to shape-radiance ambiguity. Additional depth supervision has been found to be useful in this scenario. DS-NeRF~\cite{deng2022depth} firstly demonstrates the efficacy of depth information for NeRF with sparse inputs using the coarse point clouds from structure-from-motion.
Dense depth priors~\cite{roessle2022dense} train an additional network for depth completion and uncertainty estimation, demonstrating the effectiveness of dense depth supervision. For street views, sparse LiDAR observations can be incorporated to supervise NeRF's geometry~\cite{rematas2022urban, xie2023s, wang2023neural}. Apart from using ground truth depth, existing works also show that estimated monocular depth is also helpful in improving neural 3D reconstruction and view synthesis. 
MonoSDF~\cite{yu2022monosdf} find monocular depth cues with off-the-shelf predictors can improve the quality and optimization time of neural surface reconstruction. NICER-SLAM~\cite{zhu2023nicer} also integrates monocular depth to facilitate the mapping process in SLAM for indoor scenes.
With the help of point clouds from monocular depth estimation,
NoPe-NeRF~\cite{bian2022nope} and Meuleman et al.~\cite{meuleman2023progressively} reconstruct NeRF and jointly estimate camera poses from a sequence of frames.

\topic{Depth Recovery.}
Current depth recovery technology can be roughly classified into three categories: (1) Depth completion. The goal of depth completion is to recover a dense depth map from a sparse one, e.g., the depth acquired from LiDAR. Depth completion is divided into unguided methods and guided methods. Unguided methods~\cite{uhrig2017sparsity, yang2019hierarchical, eldesokey2020uncertainty} aim at directly completing a sparse depth map with a deep neural network. Guided methods~\cite{xu2019depth, liu2021fcfr, khan2021sparse} use RGB images to complete sparse depth maps to dense. Several strategies are proposed to improve the performance of depth completion, such as early fusion~\cite{ma2019self, qu2020depth}, late fusion~\cite{tang2020learning, liu2023mff}, residual depth models~\cite{liao2017parse, liu2021learning}, and spatial propagation network based networks~\cite{cheng2018depth, park2020non}. (2) Monocular depth estimation. The goal of monocular depth estimation is to estimate a depth map from a single image. Early work mainly employs handcrafted feature \cite{handfeature1, mrf} to do monocular depth estimation, which often fails in complicated scenes. Currently, learning-based methods have shown their superiority, and encoder-decoder networks \cite{li2020enhancing, cao2021learning, chen2021aggnet, shen2021learning} are the most commonly used architecture in this area. (3) Binocular depth estimation. The goal of binocular depth estimation, \ie stereo matching is to estimate a disparity/depth map from a pair of stereo images. It is a classic task, and a well-known four-step pipeline~\cite{scharstein2002taxonomy} has been established. The early learning-based method~\cite{mccnn, luo} mainly employs a convolutional network to replace one step of the traditional pipeline, \ie feature extraction. GCNet~\cite{gcnet} is a breakthrough, which first proposes an end-to-end network to mimic the steps of the typical pipeline. Then, better feature extraction \cite{mcvmfc, emcua}, cost volume construction~\cite{gwcnet, cascade}, cost aggregation \cite{ganet, aanet, psmnet}, and disparity computation \cite{acfnet, pds} are proposed by the follow-up methods to optimize the pipeline further. 
\section{Depth-supervised NeRF}
\label{sec:preliminary}

As mentioned in \secref{sec:intro}, prior work has proved that depth priors are beneficial for NeRF training, especially in outdoor scenes, and multiple methods have been proposed to merge the depth prior into the NeRF framework. \tabref{tab:nerfdepth} classifies the existing depth-supervised NeRF methods, and two observations can be concluded from this table: 
\begin{itemize}
    \item Multiple depth priors have been applied in depth-supervised NeRF methods. However, all these methods only test one kind of depth prior and do not compare with other ones. Hence, the criteria for selecting depth priors and the relative merits of different priors have not been thoroughly investigated. Moreover, one available depth prior, binocular depth estimation, is missed in all existing work. 
    \item Existing depth-supervised NeRF methods have proposed multiple loss functions to merge the depth prior into the NeRF framework. Similar to the former one, the relative merits of selecting different loss functions to use the depth priors is also an unexplored problem. 
\end{itemize}
Hence, it is necessary to provide a comprehensive study and evaluation of employing depth priors to outdoor neural radiance fields. Specifically, we will give a classification of current depth-supervised NeRF methods and employed depth priors in this section.

\subsection{Taxonomy of Depth-supervised NeRF}
NeRF~\cite{mildenhall2020nerf} represents a 3D scene as a continuous function that maps 3D positions $\textbf{x} \in \mathbb{R}^{3}$ and 2D view directions $\textbf{d} \in \mathbb{R}^{2}$ to radiance colors $\textbf{c} \in \mathbb{R}^{3}$ and densities $\theta \in \mathbb{R}$. The function is typically parametrized with a MLP $f_\theta: (\textbf{x}, \textbf{d}) \rightarrow (\textbf{c}, \theta)$. To render an image $\mathbf{I}$, we integrate the color along each camera ray $\mathbf{r}(t) = \mathbf{o} + t\mathbf{d}$ that shots from the camera center $\mathbf{o}$ in direction $\mathbf{d}$ with volume rendering:
\begin{equation}
    \mathbf{I}_{\theta}(\mathbf{r}) = \int_{t_n}^{t_f}T(t)\sigma(\mathbf{r}(t))\mathbf{c}(\mathbf{r}(t), \mathbf{d}) \mathrm{d}t,
\label{equ:rgb_render}
\end{equation}
where $T(t) = \mathrm{exp}(-\int_{t_n}^{t}\sigma(\mathbf{r}(t))\mathrm{d}t)$ denotes the accumulated transmittance indicating the probability that a ray travels from $t_n$ to $t$ without hitting any particle. Similar to color, the depth map in NeRF can be rendered as follows:
\begin{equation}
    \mathbf{D}_{\theta}(\mathbf{r}) = \int_{t_n}^{t_f}T(t)\sigma(\mathbf{r}(t))t \mathrm{d}t.
\label{equ:depth_render}
\end{equation}

\begin{table}[t!]
\centering
\caption{Classification of existing depth-supervised NeRF methods.}
\vspace{-\baselineskip}
\begin{tabular}{ccc}
\toprule
Methods & Depth priors & Loss Type \\ 
\midrule
DS-NeRF~\cite{deng2022depth} & SfM & KL    \\ 
Urban-NeRF~\cite{rematas2022urban} & LiDAR & URF \\
S-NeRF~\cite{xie2023s} & Completion & L1 \\
MonoSDF~\cite{yu2022monosdf} & Mono & MSE \\
NICER-SLAM~\cite{zhu2023nicer} & Mono & MSE \\
NoPe-NeRF~\cite{bian2022nope} & Mono & L1 \\
FEGR~\cite{wang2023neural} & LiDAR & L1 \\
\bottomrule
\end{tabular}
\label{tab:nerfdepth}
\end{table}

\begin{table}[t!]
\centering
\caption{Taxonomy of depth priors. ($\cdot$) denotes an optional selection.}
\vspace{-\baselineskip}
\resizebox{0.48\textwidth}{!}{
\begin{tabular}{lcccc}
\toprule
Depth priors                    & LiDAR  & Camera & Cost  & Density \\ \midrule

Raw LiDAR depth                      & \ding{52}                  & \ding{56}           & high  & sparse \\ 
Depth completion           & \ding{52}                  & (\ding{52})        & high  & dense \\ 
Monocular depth estimation & \ding{56}                   & \ding{52}          & low    & dense \\ 
Binocular depth estimation & \ding{56}                   & \ding{52}          & middle  & dense \\ \toprule
\end{tabular}
}
\label{tab:Taxonomy of depth priors}
\end{table}

Given a set of posed images $\mathcal{I} = \{I_i | i = 0, 1, ..., N\}$, vanilla NeRF is optimized by comparing the mean-square error (MSE) between rendered images and their ground truth: $\mathcal{L}_{\mathrm{MSE}}^{\mathrm{rgb}} = \sum_{i}^{N} ||I_{i} - \hat I_{i}||_{2}^{2}$. Follow-up works augment NeRF training with additional depth information. To perform depth supervision, NeRF-based methods first sample a batch of $N_{r}$ rays, then compare the rendered depth and ground truth depth with different loss functions. Existing NeRF-based methods use depth supervision from two aspects: direct or indirect. Below we will introduce each category for more details. 

\topic{Direct supervision.} Direct supervisions directly compare the depth rendered by NeRF with that of the depth prior using supervision loss, including MSE and L1:
\begin{align}
\mathcal{L}_{\mathrm{MSE}}^{d} &= \sum_{i}^{N_r}|| D(r_i) - \hat D(r_i) ||^{2}, \\
\mathcal{L}_{\mathrm{L1}} &= \sum_{i}^{N_r}|D(r_i) - \hat D(r_i)|,
\label{direct}
\end{align}
where $D(r_i)$ and $\hat D(r_i)$ are the predicted and ground truth depth of ray $r_i$. Both L1 and MSE loss are included in our experiment. 
 
\topic{Indirect supervision.} Indirect supervision uses depth prior to regularize the weights of NeRF, including the KL loss in DS-NeRF~\cite{deng2022depth} and URF loss in Urban-NeRF~\cite{rematas2022urban}:
\begin{align}
\mathcal{L}_{\mathrm{KL}} &= \sum_{i}^{N_r}\sum_{k}\mathrm{log}{w_{k}}\:\mathrm{exp}(-\frac{(t_k - D(r_{i}))^2}{2\hat\sigma^{2}})\Delta t_{k}, \\
\mathcal{L}_{\mathrm{URF}} &= \sum_{t=t_n}^{D(r_i)-\epsilon}w(t)^2 + \sum_{t=D(r_i)-\epsilon}^{D(r_i)+\epsilon}(w(t)-\mathcal{K}_{\epsilon}(t - D(r_i))^2,
\label{indirect}
\end{align}
where $D(r_i)$ and $\hat D(r_i)$ are the predicted and ground truth depth of ray $r_i$, $w_k$ is the rendering weights of NeRF, $t_k$ and $\Delta t_{k}$ are the sampled points and distances of ray $r_i$, $w(t)$ is the weight corresponding to the point of distance $t$, $\mathcal{\epsilon}(x)$ is a kernel that integrates to one (\ie a distribution) and has a bounded domain parameterized by $\epsilon$.
As the URF loss has no open-source implementation, we select the KL loss as the representative of indirect supervision.  

\subsection{Taxonomy of Depth Priors}
As mentioned before, raw LiDAR depth, depth completion, monocular depth estimation, and binocular depth estimation are the main depth priors employed in outdoor neural radiance fields. The taxonomy result of current depth priors is shown in \tabref{tab:Taxonomy of depth priors} and we can conclude two observations from this table:

\begin{itemize}
    \item Raw LiDAR depth and depth completion are LiDAR-based methods. One major limitation of these methods is that the employed LiDAR is cumbersome and provides only sparse measurements. On that basis, depth completion is proposed to recover a dense depth map from a sparse one.
    \item Monocular depth estimation and binocular depth estimation are camera-based methods. These methods can directly infer a dense depth from images, which is a much cheaper selection while making a large compromise in accuracy. 
\end{itemize}
Below we will introduce each method in more detail.

\topic{Raw LiDAR depth.} The raw LiDAR depth can be directly acquired from the employed optical devices, \ie LiDAR. As the LiDAR only provides sparse depth measurements, some methods also select to combine multi-frame of the point cloud \cite{kitti1, kitti2} to get a denser depth map. 

\topic{Depth completion.} 
Denote the input raw LiDAR depth as $D_l$ and the corresponding image as $I$. The depth completion process can be represented as:
\begin{equation}
\begin{aligned}
{D_{guide}}  &= \delta (I,{D_l}),\\
{D_{unguide}} &= \delta ({D_l}),
\end{aligned}
\end{equation}
where $D_{guide}$ and $D_{unguide}$ denote unguided methods and guided methods, respectively. Currently, the latter can achieve higher accuracy with guidance from the image. Note that the image information is also the necessary input for NeRF. Hence, we select the guided method MFFNet~\cite{liu2023mff} as our depth completion method.

\topic{Monocular depth estimation.} 
Encoder-decoder networks \cite{bts, lapdepth, adabins} are the most commonly used architecture for this task. 
Let us define the input image as $I$. The whole monocular depth estimation process can be represented as:
\begin{eqnarray}
{D_{mono}} = {\varphi _d}({\varphi _e}(I)),
\end{eqnarray}
where $\varphi _e$ denotes the encoder and $\varphi _d$ denotes the decoder. We select a representative encoder-decoder network BTS \cite{bts} as our monocular depth estimation method. Note that this method is supervised and the outputs have the correct scale. For self-supervised or zero-shot pre-trained monocular depth estimation models, additional scale and drift should be estimated during training.

\topic{Binocular depth estimation.} 
Feature extraction, cost volume construction, cost aggregation, and disparity computation are the typical pipeline of current deep stereo matching methods. Denote the input left and right images as $I_l$ and $I_r$. The whole binocular depth estimation process can be represented as:
\begin{eqnarray}
d = \eta (\delta (\partial ({f_\theta }({I_l}),{f_\theta }({I_r})))),
\end{eqnarray}
where $f_\theta$ is the feature extraction network, $\partial$ is the cost volume construction network, $\delta$ is the cost aggregation network, and $\eta$ is the disparity computation step. We select two representative stereo matching networks CFNet~\cite{cfnet} and PCWNet~\cite{pcwnet} as our binocular depth estimation method.

\section{Experiments and Findings}
\label{sec:results}
In this section, we introduce the experiment settings and results of this paper. More details can be found in the \href{\suppurl}{supplementary}.


\subsection{Dataset} 
KITTI~\cite{geiger2012KITTI} and Argoverse\cite{wilson2023argoverse} are large datasets of real-world outdoor driving scenes. We evaluate and compare these methods on selected fragments from the KITTI odometry and Argoverse stereo sequences.
In contrast to the object-centric datasets commonly used in NeRF, the vehicles in autonomous driving scenarios usually only move forward or turn.
To reduce the influence of lighting changes and moving objects, 
we finally select five sequences from Seq\ 00, 02, 05, 06 in KITTI (125, 133, 175, 295, 320 frames) and three sequences from Argoverse (73, 72, 73 frames). 
Please refer to the supplemental material for details.
For each sequence, we hold every 10 frames as the testing set, and the others are used for training. 
To verify the impact of sparse viewpoints, we simulate low-frequency imaging at 2.5\,Hz. To this end,
we select $25\%$ of KITTI training data, i.e., taking one for every 4. For the Argoverse dataset, we select $50\%$ of training data, since its data logging frequency~(5\,Hz) is 1/2 of that in KITTI~(10\,Hz).
We also make experiments on all the training data, i.e., dense viewpoints. For the pose of the images, to avoid the inconsistency between the structure from motion~(SfM) pose and the real depth scale, we use the poses provided by KITTI odometry and tracking poses from Argoverse.

\begin{table*}
\centering
\caption{Quantitative comparison with selected methods on the KITTI dataset. The best and the second best results are shown in \textbf{bold} and \underline{underlined} forms, respectively.}
\resizebox{0.95\textwidth}{!}{%
\begin{tabular}{llccccccccccccccc}
\toprule
&  & \multicolumn{5}{c}{Dense} & \multicolumn{5}{c}{Sparse} \\
Methods & Depth Supervision & PSNR$\uparrow$ & SSIM$\uparrow$ & LPIPS$\downarrow$ & RMSE$\downarrow$ & ABSREL$\downarrow$ & PSNR$\uparrow$ & SSIM$\uparrow$ & LPIPS$\downarrow$ & RMSE$\downarrow$ & ABSREL$\downarrow$ \\
\cmidrule(lr){1-1}
\cmidrule(lr){2-2}
\cmidrule(lr){3-7}
\cmidrule(lr){8-12}
\multirow{5}{*}{MipNeRF-360~\cite{barron2022mip}}  & RGB-Only & \textbf{21.99} & \textbf{0.692} & \textbf{0.437} & 3.090 & 0.088 & 16.93 & 0.589 & 0.498 & 4.662 & 0.144  \\
& GT Depth & \underline{21.84} & \underline{0.682} & \underline{0.451} & \underline{0.918} & 0.032 & 19.14 & 0.630 & \textbf{0.474} & \underline{1.044} & 0.040 \\
& Depth Completion & 21.51 & 0.670 & 0.467 & \textbf{0.818} & \textbf{0.026} & \underline{19.65} & \underline{0.631} & 0.482 & \textbf{1.030} & \textbf{0.032} \\
& Stereo Depth & 21.53 & 0.665 & 0.469 & 1.192 & \underline{0.030} & \textbf{19.80} & \textbf{0.637} & \underline{0.475} & 1.246 & \underline{0.034} \\
& Mono Depth & 21.48 & 0.668 & 0.468 & 2.161 & 0.059 & 19.35 & 0.625 & 0.485 & 1.890 & 0.058 \\
\midrule
\multirow{5}{*}{NeRF++~\cite{zhang2020nerf++}}  & RGB-Only & \textbf{20.29} & 0.520 &	0.585 &	48.638 & 3.917 & 17.60 & 0.535 & 0.562 & 56.253 & 4.960  \\
& GT Depth & 20.08 & 0.574 & 0.563 & \textbf{1.914} & \textbf{0.078} & \textbf{18.90} & \textbf{0.554} & \textbf{0.568} & \textbf{1.882} & \textbf{0.089} \\
& Depth Completion & \underline{20.15} & \textbf{0.576} & \textbf{0.560} & 2.618 & 0.102& \textbf{18.90} & \underline{0.553} & \underline{0.569} & \underline{2.022} & \underline{0.094} \\
& Stereo Depth & 20.10 & \underline{0.575} & \textbf{0.560} & \underline{1.934} & \underline{0.087} & 18.85 & 0.550 & 0.574 & 2.061 & 0.100 \\
& Mono Depth & 19.87 & 0.566 & 0.567 & 2.256 & 0.092 & 18.74 & 0.548 & 0.574 & 2.670 & 0.110 \\
\midrule
\multirow{5}{*}{Instant-NGP~\cite{muller2022instant}}  
& RGB-Only & 20.51 & 0.630 & \underline{0.460} & 9.575 & 0.507
           & 15.44 & 0.499 & 0.536 & 15.011 & 0.793 \\
& GT Depth & \textbf{21.31} & \textbf{0.650} & \textbf{0.444} & \textbf{1.571} & \underline{0.052} 
           & 18.53 & \textbf{0.586} & \textbf{0.469} & \textbf{1.751} & \underline{0.060} \\
& Depth Completion & 20.90 & \underline{0.632} & 0.470 & \underline{1.661} & \textbf{0.050} 
                   & \textbf{18.62}	& \underline{0.576} & \underline{0.492} & \underline{1.833} & \textbf{0.059} \\
& Stereo Depth & \underline{20.93} & 0.629 & 0.472 & 1.830 & 0.057 
               & \underline{18.60} & 0.574 & 0.493 & 1.984 & 0.064 \\
& Mono Depth & 20.59 & 0.617 & 0.483 & 2.679 & 0.085 
             & 18.17 & 0.557 & 0.502 & 2.868 & 0.096 \\
\bottomrule
\end{tabular}
} 

\label{tab:KITTI}
\end{table*}

\begin{table*}[!htb]
\centering
\caption{Quantitative evaluations of depth prior on the KITTI dataset with the selected sequences. Please refer to the previous work \cite{bts,monodepth2_iccv2019} for the specific definition of evaluation metrics.}
\footnotesize
\resizebox{0.95\linewidth}{!}{%
{
\begin{tabular}{lccc|cccc|c}
\toprule
\multirow{1}{*}{Methods}  & \multicolumn{1}{c}{$\delta\!<\!{1.25}\!\uparrow$} & \multicolumn{1}{c}{$\delta\!<\!{1.25^2}\!\uparrow$} & $\delta\!<\!{1.25^3}\!\uparrow$ & \multicolumn{1}{c}{ABSREL$\downarrow$} & \multicolumn{1}{c}{Sq Rel$\downarrow$} & \multicolumn{1}{c}{RMSE$\downarrow$ } & RMSE log$\downarrow$ & Density \\
 \midrule
                        
Monocular Estimation         & \multicolumn{1}{c}{0.970}            & \multicolumn{1}{c}{0.997}             & 0.999             & \multicolumn{1}{c}{0.058}   & \multicolumn{1}{c}{0.156}  & \multicolumn{1}{c}{2.020} & 0.085 & 100\%   \\
Stereo Matching                & \multicolumn{1}{c}{0.996}            & \multicolumn{1}{c}{0.998}             & 0.999             & \multicolumn{1}{c}{0.016}   & \multicolumn{1}{c}{0.035}  & \multicolumn{1}{c}{1.080} & 0.040 & 100\%   \\
Stereo Matching\_confidence                & \multicolumn{1}{c}{\textbf{0.999}}            & \multicolumn{1}{c}{0.999}             & 0.999             & \multicolumn{1}{c}{0.014}   & \multicolumn{1}{c}{0.016}  & \multicolumn{1}{c}{0.71} & 0.025 & 92.28\%    \\
Depth Completion  &\multicolumn{1}{c}{0.998} & \multicolumn{1}{c}{\textbf{1.000}} & \textbf{1.000} & \multicolumn{1}{c}{\textbf{0.010}} & \multicolumn{1}{c}{\textbf{0.015}} & \multicolumn{1}{c}{\textbf{0.622}} & \textbf{0.020} & 100\% \\ \bottomrule
\end{tabular}
}}
\label{tab: depth_result}
\end{table*}

\begin{figure*}
    \centering
    \includegraphics[width=1.0\textwidth]{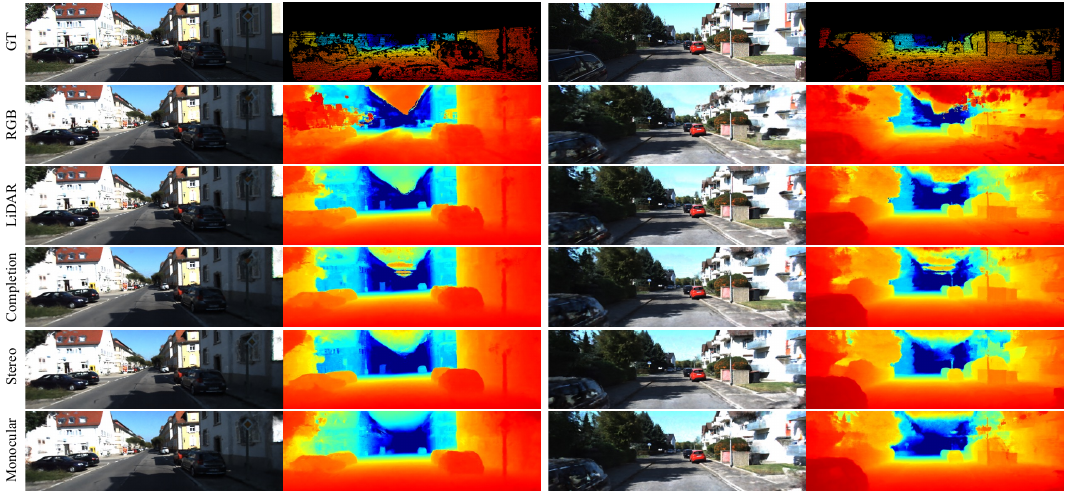}
    \caption{Qualitative results on the KITTI dataset with MipNeRF-360 with sparse viewpoints. Compared with training with RGB, adding depth supervision improves quality significantly. Better viewed zoomed-in and in-color.}
    \label{fig:KITTI_res}
\end{figure*}

\begin{figure*}
    \centering
    \includegraphics[width=0.95\textwidth]{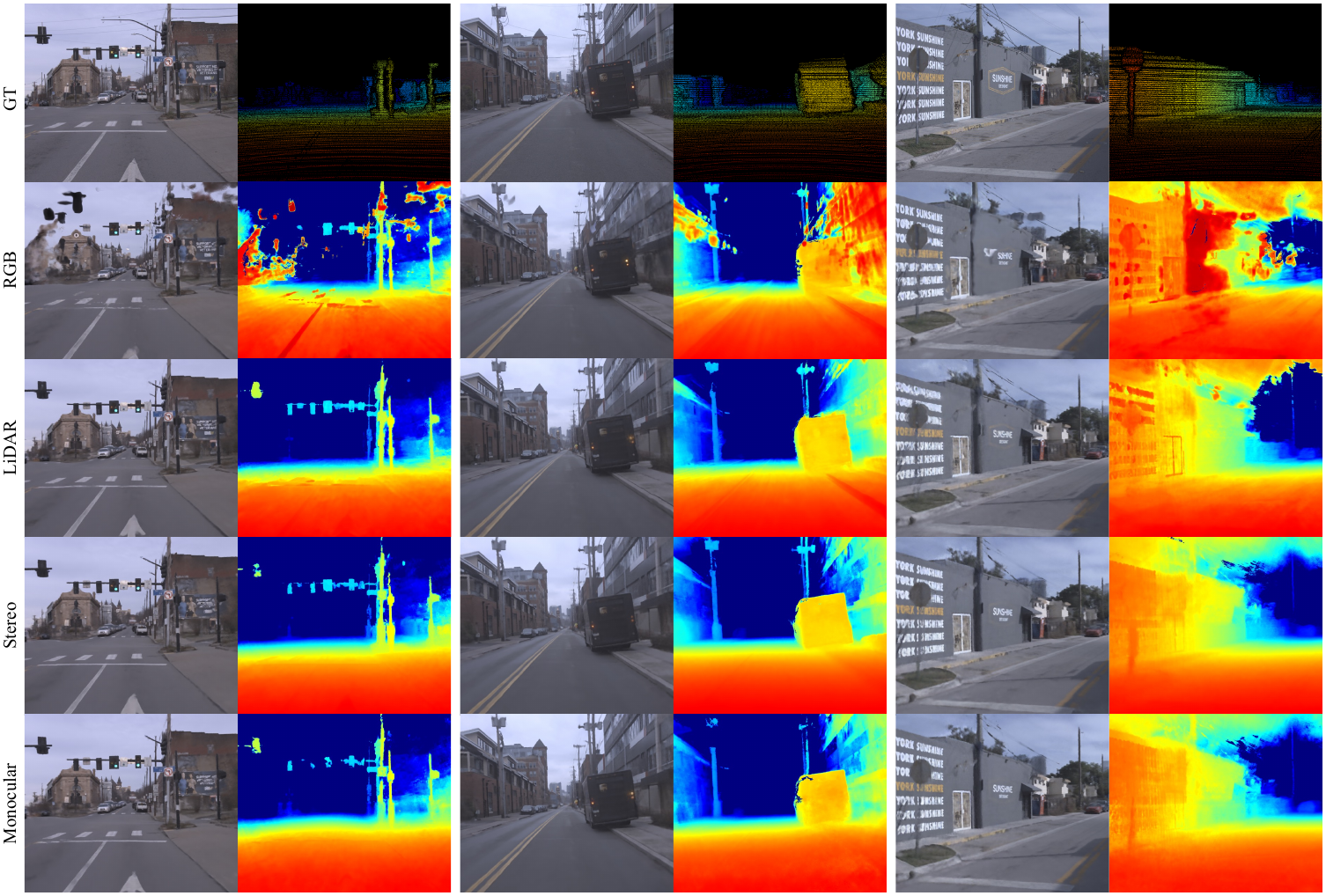}
    \caption{Qualitative results on the Argoverse dataset with MipNeRF-360 with sparse viewpoints (the GT depth is dilated with a $3\times 3$ kernel for better visualization, which is extremely sparse). Compared with training with RGB, adding depth supervision improves quality significantly. Better viewed zoomed-in and in-color.}
    \label{fig:argo_res}
\end{figure*}

\subsection{Evaluation Metrics}
\subsubsection{Photorealistic Metrics}
We use the common metrics in the novel view synthesis literature to compare the synthesized views at testing viewpoints with the ground truth: PSNR, SSIM~\cite{wang2004image} and LPIPS~\cite{zhang2018unreasonable}.

\subsubsection{Depth Accuracy Metrics} 
Following previous work~\cite{bts,monodepth2_iccv2019}, we use ABSREL (Mean Absolute Relative Error) and RMSE (Root Mean Squared Error) as our depth evaluation metrics. 


\subsection{Included NeRF Baselines}
The original parametrization of NeRF can only deal with bounded or forward-facing scenes. Since we deal with unbounded real scenes, we select the following NeRF variants in our experiments.

\topic{NeRF++}\cite{zhang2020nerf++} divides the unbounded scenes into two volumes: an inner unit sphere and an outer volume. Therefore, the volume rendering also consists of two parts. We render the depth in NeRF++ with an extended version of \equref{equ:depth_render}:
\begin{gather}
\begin{aligned}
    &\mathbf{D}(\mathbf{r}) = \int_{t = 0}^{t^{\prime}} \sigma(\mathbf{r}(t))t \cdot e^{-\int_{s=0}^{s}\sigma(\mathbf{r}(s))\mathrm{d}s} \mathrm{d}t \quad + \\
    &e^{-\int_{s=0}^{t^{\prime}}\sigma(\mathbf{r}(s))\mathrm{d}s} \cdot \int_{t = t^{\prime}}^{\infty} \sigma(\mathbf{r}(t))t \cdot e^{-\int_{s=t^{\prime}}^{s}\sigma(\mathbf{r}(s))\mathrm{d}s} \mathrm{d}t,
\end{aligned}
\end{gather}
where $t \in (0, t^{\prime})$ is inside the sphere and $t \in (t^{\prime}, \infty)$ is the unbounded area.

\topic{MipNeRF-360}~\cite{barron2022mip} is an unbounded extension of MipNeRF~\cite{barron2021mipnerf}, which proposes to use a contract function to parameterize the 3D Euclidean space within a ball of the radius of $2$. For MipNeRF-360, we can directly render images and depth maps in the contracted space with \equref{equ:rgb_render} and \equref{equ:depth_render}.


\topic{Instant-NGP}~\cite{muller2022instant} proposes a novel scene representation that bounds an actual scene into an axis-aligned bounding box and uses a small neural network augmented by a multi-resolution hash table of trainable feature vectors whose values are optimized through stochastic gradient descent. The features are further mapped to color and density.
Since it still uses standard volume rendering, the depth in Instant-NGP can be similarly rendered as in vanilla NeRF (\equref{equ:depth_render}).


\begin{table*}
\centering
\caption{Evaluation and comparison of MipNeRF-360 and Instant-NGP on the Argoverse dataset. The best and the second best results are shown in \textbf{bold} and \underline{underlined} forms, respectively.}
\resizebox{0.95\textwidth}{!}{%
\begin{tabular}{llccccccccccccccc}
\toprule
&  & \multicolumn{5}{c}{Dense} & \multicolumn{5}{c}{Sparse} \\
Methods & Depth Supervision & PSNR$\uparrow$ & SSIM$\uparrow$ & LPIPS$\downarrow$ & RMSE$\downarrow$ & ABSREL$\downarrow$ & PSNR$\uparrow$ & SSIM$\uparrow$ & LPIPS$\downarrow$ & RMSE$\downarrow$ & ABSREL$\downarrow$ \\
\cmidrule(lr){1-1}
\cmidrule(lr){2-2}
\cmidrule(lr){3-7}
\cmidrule(lr){8-12}
\multirow{4}{*}{MipNeRF-360~\cite{barron2022mip}}  & RGB-Only & \textbf{29.35} & \textbf{0.855} & \textbf{0.446} & 6.113 & 0.120 & 25.81 & 0.829 & 0.468 & 6.971 & 0.139 \\
& GT Depth & \underline{28.78} &\underline{0.846} & \underline{0.458} & \textbf{2.251} & \textbf{0.044} & \underline{28.01} & \textbf{0.840} & \textbf{0.462} & \textbf{2.443} & \textbf{0.048} \\
& Stereo Depth & 28.32 & 0.837 & 0.470 & \underline{4.271} & \underline{0.064} & 27.72 & 0.833 & 0.471 & \underline{4.310} & \underline{0.067} \\
& Mono Depth & 28.58 & 0.841 & 0.466 & 4.601 & 0.093 & \textbf{28.04} & \underline{0.836} & \underline{0.468} & 4.868 & 0.093\\

\midrule
\multirow{4}{*}{Instant-NGP~\cite{muller2022instant}}  
& RGB-Only & 28.07 & \textbf{0.847} & \textbf{0.450} & 13.478 & 0.493
           & 22.18 & 0.816 & 0.494 & 17.439 & 0.593 \\
& GT Depth & \textbf{28.92} & \textbf{0.847} & \textbf{0.450} & \textbf{1.804} & \textbf{0.045} 
           & \textbf{27.38} & \textbf{0.834} & \textbf{0.460} & \textbf{1.881} & \textbf{0.048} \\
& Stereo Depth & \underline{28.32} & 0.839 & 0.460 & \underline{5.613} & \underline{0.090} 
               & \underline{27.10} & 0.828 & \underline{0.467} & \underline{5.843} & \underline{0.097} \\
& Mono Depth & 28.31 & 0.838 & 0.466 & 6.083 & 0.122 
             & 26.97 & \underline{0.829} & 0.471 & 6.643 & 0.132 \\
\bottomrule
\end{tabular}
} 
\label{tab:argoverse}
\vspace{-0.1in}
\end{table*}

\subsection{Evaluation Results and Comparisons}
In this section, we conduct experiments on both Argoverse and KITTI datasets to verify the relative merits of employing different depth priors. Below we describe each dataset’s result in more detail.

\topic{KITTI} The qualitative depth-supervised NeRF results and corresponding depth prior quality evaluation can be found in \tabref{tab:KITTI} and \ref{tab: depth_result}. As shown in \tabref{tab: depth_result}, the performance gap between different depth priors is large. Specifically, depth completion has the highest accuracy then goes with binocular depth estimation and monocular depth estimation. Below we will further analyze the relative merits of employing different depth priors to dense and sparse views.

(1)~\textbf{Sparse view.} We first discuss the experiment result on the sparse view setting. As shown, NeRF trained with pure RGB suffers from heavy shape-radiance ambiguity (\figref{fig:KITTI_res}), so the view synthesis quality at novel viewpoints degrades significantly. In this case, the importance of depth information is evident, and we can see from \tabref{tab:KITTI} that any type of depth can be conducive and greatly improve the synthesized views. Taking MipNeRF-360 as an example, we can see 11.55\%$\sim$14.49\% photorealistic metrics improvement (PSNR) and 59.72\%$\sim$77.78\% depth accuracy metrics improvement (ABSREL) with any type of depth prior. Moreover, we can observe that the photorealistic performance gain of using different depth prior is close even if the depth quality gap between different depth priors is large. That is, we can employ the cheapest depth prior (\ie monocular depth estimation) to achieve similar performance improvement with the costly depth prior (\eg the ground truth depth collected by LiDAR). A similar situation can be observed on Instant-NGP. Hence, we can get our \textbf{finding 1: Monocular depth is enough for sparse view.} Our first counter-intuitive finding is that using monocular depth estimation can significantly improve the quality of NeRF and even achieve comparable results with the ground truth depth supervision in sparse view. Generally, the monocular depth estimation is a cheaper selection and does not need additional equipment, \eg LiDAR. Thus, monocular depth estimation is a better selection in sparse view and binocular depth estimation is also an option if you need better depth map quality.

(2)~\textbf{Dense view.} We then discuss the experiment result on the dense view setting. As shown, depth supervision is also helpful for the depth accuracy metrics. Taking MipNeRF-360 as an example, we can see 32.96\%$\sim$70.45\% depth accuracy metrics improvement (ABSREL) with any type of depth prior. That is the depth prior is still essential for the radiance field to obtain a reasonable underlying geometry. On the other hand, the performance gain in photorealistic metrics is not so noteworthy (Instant-NGP) or even causes a performance drop in some methods (MipNeRF-360). We attribute the drop in performance to inconsistent optimization directions for depth and RGB under contraction functions since one of the main differences between Instant-NGP and MipNeRF-360 is the usage of unbounded contraction. As a result, we can get our \textbf{finding 2: depth supervision is an option for dense view.} Our second interesting finding is that using depth supervision can achieve significant geometry improvement and trivial photorealistic metrics improvement in dense view. Thus, depth supervision is an option in dense view. It is still necessary if the corresponding application needs the employed NeRF to have a better geometry, such as reconstruction and potentially relighting, shadowing, \etc.

\topic{Argoverse} We also experimented on the Argoverse dataset to further verify our claim. Please note that the Argoverse dataset does not provide the depth completion task, so we do not include this setting in the experiment. The qualitative results can be found in \tabref{tab:argoverse}. As shown, the view synthesis quality at novel viewpoints also degrades significantly in sparse views, and any depth can greatly improve the synthesized views. We also observe significant depth accuracy metrics improvement and trivial photorealistic metrics for the dense view setting. The situation is the same with the KITTI dataset, which further supports the validity of our findings 1\&2.

\subsection{Ablation Study}
Although experimental results in Sec. 4.5 has demonstrated the relative merits of employing different depth priors, there are still many underlying settings that can reveal more profound findings, such as the impact of different depth densities, depth range, confidence level, depth loss function, etc. In this section, we conduct several ablation experiments on one sequence of the KITTI dataset with MipNeRF-360 and sparse view setting to further explore those factors.

\begin{figure}[t]
    \centering
    \includegraphics[width=\linewidth]{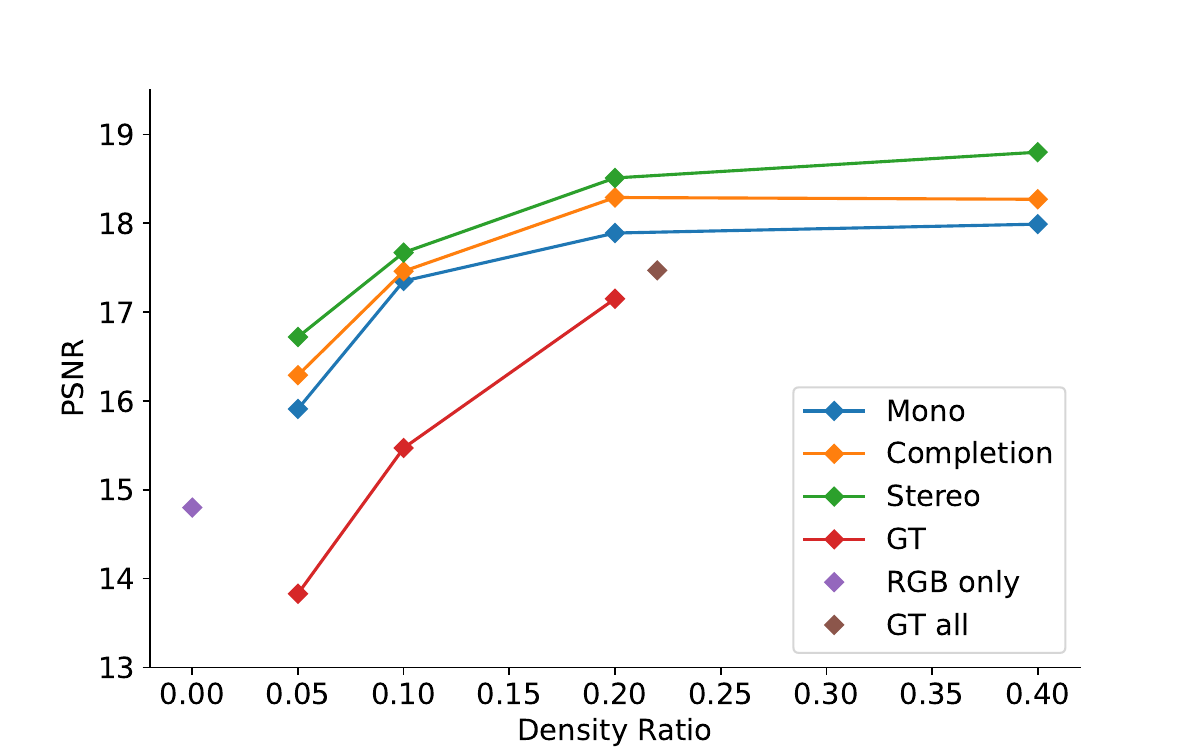}
    \caption{The relationship between PSNR and the density ratio of depth supervision under different types of depth priors.}
    \label{fig:ratio}
\end{figure}

\subsubsection{Density} 
Density is the main factor that differentiates the LiDAR ground truth from other depth priors, e.g. the density of LiDAR ground truth on the KITTI dataset is nearly $20\%$, and the other depth priors is 100\%. We conduct two ablation experiments to further investigate the influence of different densities.

\topic{GT Masking} It is interesting that monocular depth estimation can achieve on-par or even better performance than ground truth LiDAR in photorealistic metrics. This is likely because monocular depth is much denser than LiDAR, which only provides valid depth gt for around $20\%$ of pixels. To verify this hypothesis, we try to use the invalid location of ground truth LiDAR to mask out the monocular depth estimation result and make sure the two depth priors have the same density. The results show that the photorealistic performance of using monocular depth drops to be close to or inferior to using ground truth depth after masking. This indicates that even imprecise dense supervision can still be more beneficial than sparse one in sparse views for novel view synthesis.

\topic{Depth Density} To further evaluate the influence of supervision sparsity, we also test different depth densities by iteratively removing a fixed proportion of pixels from the original depth priors. The results in \figref{fig:ratio} show that a tiny amount ($5\%$) of depth supervision is enough to improve the performance of NeRF, and a denser depth supervision can achieve a more considerable gain. Specifically, the improvement is more significant before $20\%$, and the results become stable after enough supervision ($40\%$). Moreover, we also observe that even with randomly selected less than 20\% of monocular depth supervision (or depth completion and stereo depth), the results already surpass using all the ground truth (22\%), which indicates that imprecise while covering a wider range of areas supervision (e.g., monocular depth estimation) is better than accurate while limited in central regions supervision (e.g., LIDAR) in sparse view. The GT Masking experiment further supports our claim, i.e., a sparse and covering the same region monocular depth estimation cannot beat the ground truth LIDAR supervision.
 
\vspace{6pt} \noindent Considering the above experiments, we can obtain our \textbf{finding 3: The denser the better.}
Our third interesting finding is that even a very sparse depth supervision can significantly boost the quality of novel view synthesis in sparse view, and generally, we observe that the denser the depth, the better quality of novel view synthesis we can obtain.    

\subsubsection{Depth Loss Type} Apart from MSE, we also test L1 and KL losses. As shown in \tabref{tab:ablation}, there are no significant performance differences between L1 and MSE loss. Moreover, KL loss is significantly worse than MSE, possibly because it is an indirect loss function and has a strong constraint in NeRF optimization, which would cause adverse effects if the depth estimation is not accurate enough.

\subsubsection{Depth Filtering} The quality of depth varies widely among different methods. Generally, the quality degrades from completion to stereo, and finally to monocular (refer to \tabref{tab: depth_result}). To address the unstable presentation of depth quality in many downstream tasks, there exist a few approaches to filter certain depth values and only use a subset of them. In this study, we choose two simple and widely-used depth filtering approaches, i.e., threshold clipping and confidence-based filtering, to investigate the influence.

\topic{Threshold Clipping} In \tabref{tab:argoverse} and \tabref{tab:KITTI}, we use depth prediction from different methods directly (cropping the sky area). Generally, the background area (far location) is less accurate than the foreground area (close location). Here, we test the impact of filtering out predictions farther than the threshold, \ie large than a threshold $s$. In our experiments, we set $s$ as $40m$ and $80m$. The results are shown in \tabref{tab:ablation}. We can see that there are no significant differences in both image and depth metrics before and after clipping with $s=80m$. For $s=40m$, the image and depth metrics drop a little, likely because it lost the depth information of the background area, although it is less accurate than the foreground area.

\topic{Confidence-based Filtering} Confidence estimation is also a common operation in both depth estimation and depth-supervised NeRF. Here, we test such confidence filters in binocular depth estimation as an example. Specifically, the confidence filters are implemented by the uncertainty estimation proposed in CFNet. The corresponding depth quality evaluation result is shown in \tabref{tab: depth_result}. From \tabref{tab:ablation}, we can see that similar to threshold clipping, after filtering, the performance has no notable change.

\vspace{3.5pt}
\noindent \textbf{Finding 4: Simple loss function and depth filtering are enough.}
The above three ablation studies lead to our fourth finding: Complex depth filtering and loss function is unnecessary in outdoor NeRF and directly cropping the sky area (point at infinity) with MSE supervision is enough.

\subsection{Discussion}
Similar to DS-NeRF~\cite{deng2022depth}, we also conduct experiments by training Instant-NGP for 30 epochs and found that additional depth \textbf{accelerates convergence speed} and is also \textbf{beneficial in the extreme few-shot setting}. 
In the sparse setting, training with depth surpasses the training with only RGB with 30 epochs, even at the first epoch.
For extremely sparse views, we further select 1/8, 1/16 of the input views for training in the Kitti dataset. The resulting PSNR/LPIPS is 13.1/0.57 (RGB Only), 16.4/0.52 (Monocular depth) for 1/8, and 11.6/0.61 (RGB Only), 13.0/0.57 (Monocular depth) for 1/16. Using other depth supervision is even better than using monocular depth.

\begin{table}
\centering
\setlength{\tabcolsep}{2pt} 
\caption{Ablation study with GT masking, threshold clipping, confidence filtering, and loss function.}
\resizebox{\linewidth}{!}{%
\begin{tabular}{cccccccc}
\toprule
Experiments & Factor & Depth Type & PSNR$\uparrow$ & SSIM$\uparrow$ & LPIPS$\downarrow$ & RMSE$\downarrow$ & ABSREL$\downarrow$ \\ \midrule
- & RGB-Only & - & 14.80 & 0.475 & 0.551 & 4.569 & 0.153 \\ \midrule
\multirow{3}{*}{\shortstack{GT \\ Masking}} & - & LiDAR &  17.47 & \textbf{0.542} & \textbf{0.507} & \textbf{1.173} & \textbf{0.045} \\
 & Yes & Mono & 17.11 & 0.535 & 0.512 & 2.345 & 0.076 \\
 & No & Mono & \textbf{17.97} & \textbf{0.542} & 0.510 & 2.383 & 0.073 \\ \midrule
\multirow{3}{*}{\shortstack{Threshold\\Clipping}} & - & Mono & 17.97 & \textbf{0.542} & 0.510 & \textbf{2.383} & 0.073 \\
 & 40m & Mono & 17.60 & 0.541 & \textbf{0.509} & 2.470 & \textbf{0.071} \\
 & 80m & Mono & \textbf{18.18} & \textbf{0.542} & 0.510 & 2.390 & 0.073 \\ \midrule
\multirow{2}{*}{\shortstack{Confidence\\Filtering}} & No & Stereo & \textbf{18.87} & 0.562 & 0.501 & \textbf{1.349} & \textbf{0.0405} \\
 & Yes & Stereo & 18.85 & \textbf{0.565} & \textbf{0.495} & 1.467 & 0.0424 \\ \midrule
\multirow{3}{*}{\shortstack{Loss\\Function}} & MSE & Mono & \textbf{17.97} & \textbf{0.542} & \textbf{0.510} & \textbf{2.383} & \textbf{0.073} \\
 & L1 & Mono & 17.91 & 0.519 & 0.550 & 2.543 & 0.075 \\
 & KL & Mono & 16.55 & 0.526 & 0.515 & 2.487 & 0.076 \\
\bottomrule
\end{tabular} 
}

\label{tab:ablation}
\end{table}

\section{Conclusion}
\label{sec:conclusion}

This paper presents the first in-depth study and evaluation of employing depth priors to outdoor neural radiance fields, covering all common depth sensing technologies and most application ways. As a result, we conclude the experimental results and have interesting findings as follows: (1) \textbf{Density}: Even a very sparse depth supervision can significantly boost the view synthesis quality, and generally, the denser, the better; (2) \textbf{Quality}: (a) Monocular depth is enough for the sparse view, which can even achieve comparable results with the ground truth depth supervision. (b) depth supervision is an option for dense view, \ie the depth supervision is necessary if the corresponding application needs the employed NeRF to have a better geometry; (3) \textbf{Supervision}: Complex depth filtering and loss function is unnecessary in outdoor NeRF and directly cropping the sky area with MSE supervision is enough. We believe these findings can potentially benefit practitioners and researchers in training their NeRF models with depth priors.

\begin{acks}
This work was supported in part by the National Natural Science Foundation of China under Grants 62106258 and 62271410.
\end{acks}

\bibliographystyle{ACM-Reference-Format} 
\bibliography{iros_abrv}


\begin{thebibliography}{77}


\ifx \showCODEN    \undefined \def \showCODEN     #1{\unskip}     \fi
\ifx \showDOI      \undefined \def \showDOI       #1{#1}\fi
\ifx \showISBNx    \undefined \def \showISBNx     #1{\unskip}     \fi
\ifx \showISBNxiii \undefined \def \showISBNxiii  #1{\unskip}     \fi
\ifx \showISSN     \undefined \def \showISSN      #1{\unskip}     \fi
\ifx \showLCCN     \undefined \def \showLCCN      #1{\unskip}     \fi
\ifx \shownote     \undefined \def \shownote      #1{#1}          \fi
\ifx \showarticletitle \undefined \def \showarticletitle #1{#1}   \fi
\ifx \showURL      \undefined \def \showURL       {\relax}        \fi
\providecommand\bibfield[2]{#2}
\providecommand\bibinfo[2]{#2}
\providecommand\natexlab[1]{#1}
\providecommand\showeprint[2][]{arXiv:#2}

\bibitem[Barron et~al\mbox{.}(2021)]%
        {barron2021mipnerf}
\bibfield{author}{\bibinfo{person}{Jonathan~T Barron}, \bibinfo{person}{Ben
  Mildenhall}, \bibinfo{person}{Matthew Tancik}, \bibinfo{person}{Peter
  Hedman}, \bibinfo{person}{Ricardo Martin-Brualla}, {and}
  \bibinfo{person}{Pratul~P Srinivasan}.} \bibinfo{year}{2021}\natexlab{}.
\newblock \showarticletitle{Mip-nerf: A multiscale representation for
  anti-aliasing neural radiance fields}. In \bibinfo{booktitle}{\emph{Proc.~of
  the IEEE/CVF Intl.~Conf.~on Computer Vision (ICCV)}}.
\newblock


\bibitem[Barron et~al\mbox{.}(2022)]%
        {barron2022mip}
\bibfield{author}{\bibinfo{person}{Jonathan~T Barron}, \bibinfo{person}{Ben
  Mildenhall}, \bibinfo{person}{Dor Verbin}, \bibinfo{person}{Pratul~P
  Srinivasan}, {and} \bibinfo{person}{Peter Hedman}.}
  \bibinfo{year}{2022}\natexlab{}.
\newblock \showarticletitle{Mip-nerf 360: Unbounded anti-aliased neural
  radiance fields}. In \bibinfo{booktitle}{\emph{Proc. IEEE Conf. Comput. Vis.
  Pattern Recognit. (CVPR)}}.
\newblock


\bibitem[Bhat et~al\mbox{.}(2021)]%
        {adabins}
\bibfield{author}{\bibinfo{person}{Shariq~Farooq Bhat},
  \bibinfo{person}{Ibraheem Alhashim}, {and} \bibinfo{person}{Peter Wonka}.}
  \bibinfo{year}{2021}\natexlab{}.
\newblock \showarticletitle{Adabins: Depth estimation using adaptive bins}. In
  \bibinfo{booktitle}{\emph{Proc. IEEE Conf. Comput. Vis. Pattern Recognit.
  (CVPR)}}. \bibinfo{pages}{4009--4018}.
\newblock


\bibitem[Bian et~al\mbox{.}(2022)]%
        {bian2022nope}
\bibfield{author}{\bibinfo{person}{Wenjing Bian}, \bibinfo{person}{Zirui Wang},
  \bibinfo{person}{Kejie Li}, \bibinfo{person}{Jia-Wang Bian}, {and}
  \bibinfo{person}{Victor~Adrian Prisacariu}.} \bibinfo{year}{2022}\natexlab{}.
\newblock \showarticletitle{NoPe-NeRF: Optimising Neural Radiance Field with No
  Pose Prior}.
\newblock \bibinfo{journal}{\emph{arXiv preprint arXiv:2212.07388}}
  (\bibinfo{year}{2022}).
\newblock


\bibitem[Cao et~al\mbox{.}(2021)]%
        {cao2021learning}
\bibfield{author}{\bibinfo{person}{Yuanzhouhan Cao}, \bibinfo{person}{Yidong
  Li}, \bibinfo{person}{Haokui Zhang}, \bibinfo{person}{Chao Ren}, {and}
  \bibinfo{person}{Yifan Liu}.} \bibinfo{year}{2021}\natexlab{}.
\newblock \showarticletitle{Learning Structure Affinity for Video Depth
  Estimation}. In \bibinfo{booktitle}{\emph{Proceedings of the 29th ACM
  International Conference on Multimedia}}. \bibinfo{pages}{190--198}.
\newblock


\bibitem[Chang and Chen(2018)]%
        {psmnet}
\bibfield{author}{\bibinfo{person}{Jia-Ren Chang} {and}
  \bibinfo{person}{Yong-Sheng Chen}.} \bibinfo{year}{2018}\natexlab{}.
\newblock \showarticletitle{Pyramid stereo matching network}. In
  \bibinfo{booktitle}{\emph{IEEE Conference on Computer Vision and Pattern
  Recognition (CVPR)}}. \bibinfo{pages}{5410--5418}.
\newblock


\bibitem[Chang et~al\mbox{.}(2019)]%
        {wilson2023argoverse}
\bibfield{author}{\bibinfo{person}{Ming-Fang Chang}, \bibinfo{person}{John
  Lambert}, \bibinfo{person}{Patsorn Sangkloy}, \bibinfo{person}{Jagjeet
  Singh}, \bibinfo{person}{Slawomir Bak}, \bibinfo{person}{Andrew Hartnett},
  \bibinfo{person}{De Wang}, \bibinfo{person}{Peter Carr},
  \bibinfo{person}{Simon Lucey}, \bibinfo{person}{Deva Ramanan},
  {et~al\mbox{.}}} \bibinfo{year}{2019}\natexlab{}.
\newblock \showarticletitle{Argoverse: 3d tracking and forecasting with rich
  maps}. In \bibinfo{booktitle}{\emph{Proc. IEEE Conf. Comput. Vis. Pattern
  Recognit. (CVPR)}}. \bibinfo{pages}{8748--8757}.
\newblock


\bibitem[Chen et~al\mbox{.}(2021)]%
        {chen2021aggnet}
\bibfield{author}{\bibinfo{person}{Zhi Chen}, \bibinfo{person}{Xiaoqing Ye},
  \bibinfo{person}{Liang Du}, \bibinfo{person}{Wei Yang},
  \bibinfo{person}{Liusheng Huang}, \bibinfo{person}{Xiao Tan},
  \bibinfo{person}{Zhenbo Shi}, \bibinfo{person}{Fumin Shen}, {and}
  \bibinfo{person}{Errui Ding}.} \bibinfo{year}{2021}\natexlab{}.
\newblock \showarticletitle{AggNet for Self-supervised Monocular Depth
  Estimation: Go An Aggressive Step Furthe}. In
  \bibinfo{booktitle}{\emph{Proceedings of the 29th ACM International
  Conference on Multimedia}}. \bibinfo{pages}{1526--1534}.
\newblock


\bibitem[Cheng et~al\mbox{.}(2018)]%
        {cheng2018depth}
\bibfield{author}{\bibinfo{person}{Xinjing Cheng}, \bibinfo{person}{Peng Wang},
  {and} \bibinfo{person}{Ruigang Yang}.} \bibinfo{year}{2018}\natexlab{}.
\newblock \showarticletitle{Depth estimation via affinity learned with
  convolutional spatial propagation network}. In
  \bibinfo{booktitle}{\emph{Proc. Eur. Conf. Comput. Vis. (ECCV)}}.
  \bibinfo{pages}{103--119}.
\newblock


\bibitem[Deng et~al\mbox{.}(2022)]%
        {deng2022depth}
\bibfield{author}{\bibinfo{person}{Kangle Deng}, \bibinfo{person}{Andrew Liu},
  \bibinfo{person}{Jun-Yan Zhu}, {and} \bibinfo{person}{Deva Ramanan}.}
  \bibinfo{year}{2022}\natexlab{}.
\newblock \showarticletitle{Depth-supervised nerf: Fewer views and faster
  training for free}. In \bibinfo{booktitle}{\emph{Proc. IEEE Conf. Comput.
  Vis. Pattern Recognit. (CVPR)}}.
\newblock


\bibitem[Eldesokey et~al\mbox{.}(2020)]%
        {eldesokey2020uncertainty}
\bibfield{author}{\bibinfo{person}{Abdelrahman Eldesokey},
  \bibinfo{person}{Michael Felsberg}, \bibinfo{person}{Karl Holmquist}, {and}
  \bibinfo{person}{Michael Persson}.} \bibinfo{year}{2020}\natexlab{}.
\newblock \showarticletitle{Uncertainty-aware cnns for depth completion:
  Uncertainty from beginning to end}. In \bibinfo{booktitle}{\emph{Proc. IEEE
  Conf. Comput. Vis. Pattern Recognit. (CVPR)}}. \bibinfo{pages}{12014--12023}.
\newblock


\bibitem[Fridovich-Keil et~al\mbox{.}(2022)]%
        {yu2021plenoxels}
\bibfield{author}{\bibinfo{person}{Sara Fridovich-Keil}, \bibinfo{person}{Alex
  Yu}, \bibinfo{person}{Matthew Tancik}, \bibinfo{person}{Qinhong Chen},
  \bibinfo{person}{Benjamin Recht}, {and} \bibinfo{person}{Angjoo Kanazawa}.}
  \bibinfo{year}{2022}\natexlab{}.
\newblock \showarticletitle{Plenoxels: Radiance Fields without Neural
  Networks}. In \bibinfo{booktitle}{\emph{Proc. IEEE Conf. Comput. Vis. Pattern
  Recognit. (CVPR)}}.
\newblock


\bibitem[Geiger et~al\mbox{.}(2012a)]%
        {geiger2012KITTI}
\bibfield{author}{\bibinfo{person}{Andreas Geiger}, \bibinfo{person}{Philip
  Lenz}, {and} \bibinfo{person}{Raquel Urtasun}.}
  \bibinfo{year}{2012}\natexlab{a}.
\newblock \showarticletitle{Are we ready for autonomous driving? the kitti
  vision benchmark suite}. In \bibinfo{booktitle}{\emph{Proc. IEEE Conf.
  Comput. Vis. Pattern Recognit. (CVPR)}}.
\newblock


\bibitem[Geiger et~al\mbox{.}(2012b)]%
        {kitti1}
\bibfield{author}{\bibinfo{person}{Andreas Geiger}, \bibinfo{person}{Philip
  Lenz}, {and} \bibinfo{person}{Raquel Urtasun}.}
  \bibinfo{year}{2012}\natexlab{b}.
\newblock \showarticletitle{Are we ready for autonomous driving? the kitti
  vision benchmark suite}. In \bibinfo{booktitle}{\emph{IEEE Conference on
  Computer Vision and Pattern Recognition (CVPR)}}.
  \bibinfo{pages}{3354--3361}.
\newblock


\bibitem[Godard et~al\mbox{.}(2019)]%
        {monodepth2_iccv2019}
\bibfield{author}{\bibinfo{person}{Cl{\'e}ment Godard}, \bibinfo{person}{Oisin
  Mac~Aodha}, \bibinfo{person}{Michael Firman}, {and}
  \bibinfo{person}{Gabriel~J Brostow}.} \bibinfo{year}{2019}\natexlab{}.
\newblock \showarticletitle{Digging into self-supervised monocular depth
  estimation}. In \bibinfo{booktitle}{\emph{Proceedings of the IEEE
  International Conference on Computer Vision}}. \bibinfo{pages}{3828--3838}.
\newblock


\bibitem[Gu et~al\mbox{.}(2020)]%
        {cascade}
\bibfield{author}{\bibinfo{person}{Xiaodong Gu}, \bibinfo{person}{Zhiwen Fan},
  \bibinfo{person}{Siyu Zhu}, \bibinfo{person}{Zuozhuo Dai},
  \bibinfo{person}{Feitong Tan}, {and} \bibinfo{person}{Ping Tan}.}
  \bibinfo{year}{2020}\natexlab{}.
\newblock \showarticletitle{Cascade cost volume for high-resolution multi-view
  stereo and stereo matching}. In \bibinfo{booktitle}{\emph{IEEE Conference on
  Computer Vision and Pattern Recognition (CVPR)}}.
  \bibinfo{pages}{2495--2504}.
\newblock


\bibitem[Guo et~al\mbox{.}(2019)]%
        {gwcnet}
\bibfield{author}{\bibinfo{person}{Xiaoyang Guo}, \bibinfo{person}{Kai Yang},
  \bibinfo{person}{Wukui Yang}, \bibinfo{person}{Xiaogang Wang}, {and}
  \bibinfo{person}{Hongsheng Li}.} \bibinfo{year}{2019}\natexlab{}.
\newblock \showarticletitle{Group-wise correlation stereo network}. In
  \bibinfo{booktitle}{\emph{IEEE Conference on Computer Vision and Pattern
  Recognition (CVPR)}}. \bibinfo{pages}{3273--3282}.
\newblock


\bibitem[Herrera et~al\mbox{.}(2018)]%
        {handfeature1}
\bibfield{author}{\bibinfo{person}{Jose~L Herrera}, \bibinfo{person}{Carlos~R
  Del-Blanco}, {and} \bibinfo{person}{Narciso Garcia}.}
  \bibinfo{year}{2018}\natexlab{}.
\newblock \showarticletitle{Automatic depth extraction from 2D images using a
  cluster-based learning framework}.
\newblock \bibinfo{journal}{\emph{IEEE Trans.~on Image Processing (TIP)}}
  \bibinfo{volume}{27}, \bibinfo{number}{7} (\bibinfo{year}{2018}),
  \bibinfo{pages}{3288--3299}.
\newblock


\bibitem[Kajiya and Von~Herzen(1984)]%
        {kajiya1984ray}
\bibfield{author}{\bibinfo{person}{James~T Kajiya} {and}
  \bibinfo{person}{Brian~P Von~Herzen}.} \bibinfo{year}{1984}\natexlab{}.
\newblock \showarticletitle{Ray tracing volume densities}.
\newblock \bibinfo{journal}{\emph{ACM SIGGRAPH computer graphics}}
  \bibinfo{volume}{18}, \bibinfo{number}{3} (\bibinfo{year}{1984}),
  \bibinfo{pages}{165--174}.
\newblock


\bibitem[Kendall et~al\mbox{.}(2017)]%
        {gcnet}
\bibfield{author}{\bibinfo{person}{Alex Kendall}, \bibinfo{person}{Hayk
  Martirosyan}, \bibinfo{person}{Saumitro Dasgupta}, \bibinfo{person}{Peter
  Henry}, \bibinfo{person}{Ryan Kennedy}, \bibinfo{person}{Abraham Bachrach},
  {and} \bibinfo{person}{Adam Bry}.} \bibinfo{year}{2017}\natexlab{}.
\newblock \showarticletitle{End-to-end learning of geometry and context for
  deep stereo regression}. In \bibinfo{booktitle}{\emph{IEEE International
  Conference on Computer Vision (ICCV)}}. \bibinfo{pages}{66--75}.
\newblock


\bibitem[Khan et~al\mbox{.}(2021)]%
        {khan2021sparse}
\bibfield{author}{\bibinfo{person}{Md~Fahim~Faysal Khan},
  \bibinfo{person}{Nelson~Daniel Troncoso~Aldas}, \bibinfo{person}{Abhishek
  Kumar}, \bibinfo{person}{Siddharth Advani}, {and}
  \bibinfo{person}{Vijaykrishnan Narayanan}.} \bibinfo{year}{2021}\natexlab{}.
\newblock \showarticletitle{Sparse to dense depth completion using a generative
  adversarial network with intelligent sampling strategies}. In
  \bibinfo{booktitle}{\emph{Proceedings of the 29th ACM International
  Conference on Multimedia}}. \bibinfo{pages}{5528--5536}.
\newblock


\bibitem[Lee et~al\mbox{.}(2019)]%
        {bts}
\bibfield{author}{\bibinfo{person}{Jin~Han Lee}, \bibinfo{person}{Myung-Kyu
  Han}, \bibinfo{person}{Dong~Wook Ko}, {and} \bibinfo{person}{Il~Hong Suh}.}
  \bibinfo{year}{2019}\natexlab{}.
\newblock \showarticletitle{From big to small: Multi-scale local planar
  guidance for monocular depth estimation}.
\newblock \bibinfo{journal}{\emph{arXiv preprint arXiv:1907.10326}}
  (\bibinfo{year}{2019}).
\newblock


\bibitem[Li et~al\mbox{.}(2020)]%
        {li2020enhancing}
\bibfield{author}{\bibinfo{person}{Rui Li}, \bibinfo{person}{Xiantuo He},
  \bibinfo{person}{Yu Zhu}, \bibinfo{person}{Xianjun Li},
  \bibinfo{person}{Jinqiu Sun}, {and} \bibinfo{person}{Yanning Zhang}.}
  \bibinfo{year}{2020}\natexlab{}.
\newblock \showarticletitle{Enhancing self-supervised monocular depth
  estimation via incorporating robust constraints}. In
  \bibinfo{booktitle}{\emph{Proceedings of the 28th ACM International
  Conference on Multimedia}}. \bibinfo{pages}{3108--3117}.
\newblock


\bibitem[Liang et~al\mbox{.}(2019)]%
        {mcvmfc}
\bibfield{author}{\bibinfo{person}{Zhengfa Liang}, \bibinfo{person}{Yulan Guo},
  \bibinfo{person}{Yiliu Feng}, \bibinfo{person}{Wei Chen},
  \bibinfo{person}{Linbo Qiao}, \bibinfo{person}{Li Zhou},
  \bibinfo{person}{Jianfeng Zhang}, {and} \bibinfo{person}{Hengzhu Liu}.}
  \bibinfo{year}{2019}\natexlab{}.
\newblock \showarticletitle{Stereo matching using multi-level cost volume and
  multi-scale feature constancy}. In \bibinfo{booktitle}{\emph{IEEE
  transactions on pattern analysis and machine intelligence (TPAMI)}}.
\newblock


\bibitem[Liao et~al\mbox{.}(2017)]%
        {liao2017parse}
\bibfield{author}{\bibinfo{person}{Yiyi Liao}, \bibinfo{person}{Lichao Huang},
  \bibinfo{person}{Yue Wang}, \bibinfo{person}{Sarath Kodagoda},
  \bibinfo{person}{Yinan Yu}, {and} \bibinfo{person}{Yong Liu}.}
  \bibinfo{year}{2017}\natexlab{}.
\newblock \showarticletitle{Parse geometry from a line: Monocular depth
  estimation with partial laser observation}. In
  \bibinfo{booktitle}{\emph{Proc. IEEE Int. Conf. Robot. Automat. (ICRA)}}.
  \bibinfo{pages}{5059--5066}.
\newblock


\bibitem[Lin et~al\mbox{.}(2021)]%
        {lin2021barf}
\bibfield{author}{\bibinfo{person}{Chen-Hsuan Lin}, \bibinfo{person}{Wei-Chiu
  Ma}, \bibinfo{person}{Antonio Torralba}, {and} \bibinfo{person}{Simon
  Lucey}.} \bibinfo{year}{2021}\natexlab{}.
\newblock \showarticletitle{Barf: Bundle-adjusting neural radiance fields}. In
  \bibinfo{booktitle}{\emph{Proceedings of the IEEE/CVF International
  Conference on Computer Vision}}. \bibinfo{pages}{5741--5751}.
\newblock


\bibitem[Liu et~al\mbox{.}(2021a)]%
        {liu2021learning}
\bibfield{author}{\bibinfo{person}{Lina Liu}, \bibinfo{person}{Yiyi Liao},
  \bibinfo{person}{Yue Wang}, \bibinfo{person}{Andreas Geiger}, {and}
  \bibinfo{person}{Yong Liu}.} \bibinfo{year}{2021}\natexlab{a}.
\newblock \showarticletitle{Learning steering kernels for guided depth
  completion}.
\newblock \bibinfo{journal}{\emph{IEEE Trans.~on Image Processing (TIP)}}
  \bibinfo{volume}{30} (\bibinfo{year}{2021}), \bibinfo{pages}{2850--2861}.
\newblock


\bibitem[Liu et~al\mbox{.}(2021b)]%
        {liu2021fcfr}
\bibfield{author}{\bibinfo{person}{Lina Liu}, \bibinfo{person}{Xibin Song},
  \bibinfo{person}{Xiaoyang Lyu}, \bibinfo{person}{Junwei Diao},
  \bibinfo{person}{Mengmeng Wang}, \bibinfo{person}{Yong Liu}, {and}
  \bibinfo{person}{Liangjun Zhang}.} \bibinfo{year}{2021}\natexlab{b}.
\newblock \showarticletitle{Fcfr-net: Feature fusion based coarse-to-fine
  residual learning for depth completion}. In
  \bibinfo{booktitle}{\emph{Proceedings of the AAAI conference on artificial
  intelligence}}, Vol.~\bibinfo{volume}{35}. \bibinfo{pages}{2136--2144}.
\newblock


\bibitem[Liu et~al\mbox{.}(2023)]%
        {liu2023mff}
\bibfield{author}{\bibinfo{person}{Lina Liu}, \bibinfo{person}{Xibin Song},
  \bibinfo{person}{Jiadai Sun}, \bibinfo{person}{Xiaoyang Lyu},
  \bibinfo{person}{Lin Li}, \bibinfo{person}{Yong Liu}, {and}
  \bibinfo{person}{Liangjun Zhang}.} \bibinfo{year}{2023}\natexlab{}.
\newblock \showarticletitle{MFF-Net: Towards Efficient Monocular Depth
  Completion with Multi-modal Feature Fusion}.
\newblock \bibinfo{journal}{\emph{IEEE Robot. Automat. Lett. (RA-L)}}
  (\bibinfo{year}{2023}).
\newblock


\bibitem[Long et~al\mbox{.}(2023)]%
        {long2022neuraludf}
\bibfield{author}{\bibinfo{person}{Xiaoxiao Long}, \bibinfo{person}{Cheng Lin},
  \bibinfo{person}{Lingjie Liu}, \bibinfo{person}{Yuan Liu},
  \bibinfo{person}{Peng Wang}, \bibinfo{person}{Christian Theobalt},
  \bibinfo{person}{Taku Komura}, {and} \bibinfo{person}{Wenping Wang}.}
  \bibinfo{year}{2023}\natexlab{}.
\newblock \showarticletitle{NeuralUDF: Learning Unsigned Distance Fields for
  Multi-view Reconstruction of Surfaces with Arbitrary Topologies}. In
  \bibinfo{booktitle}{\emph{Proc. IEEE Conf. Comput. Vis. Pattern Recognit.
  (CVPR)}}.
\newblock


\bibitem[Luo et~al\mbox{.}(2016)]%
        {luo}
\bibfield{author}{\bibinfo{person}{Wenjie Luo}, \bibinfo{person}{Alexander~G
  Schwing}, {and} \bibinfo{person}{Raquel Urtasun}.}
  \bibinfo{year}{2016}\natexlab{}.
\newblock \showarticletitle{Efficient deep learning for stereo matching}. In
  \bibinfo{booktitle}{\emph{IEEE Conference on Computer Vision and Pattern
  Recognition (CVPR)}}. \bibinfo{pages}{5695--5703}.
\newblock


\bibitem[Ma et~al\mbox{.}(2019)]%
        {ma2019self}
\bibfield{author}{\bibinfo{person}{Fangchang Ma},
  \bibinfo{person}{Guilherme~Venturelli Cavalheiro}, {and}
  \bibinfo{person}{Sertac Karaman}.} \bibinfo{year}{2019}\natexlab{}.
\newblock \showarticletitle{Self-supervised sparse-to-dense: Self-supervised
  depth completion from lidar and monocular camera}. In
  \bibinfo{booktitle}{\emph{Proc. IEEE Int. Conf. Robot. Automat. (ICRA)}}.
  \bibinfo{pages}{3288--3295}.
\newblock


\bibitem[Menze and Geiger(2015)]%
        {kitti2}
\bibfield{author}{\bibinfo{person}{Moritz Menze} {and} \bibinfo{person}{Andreas
  Geiger}.} \bibinfo{year}{2015}\natexlab{}.
\newblock \showarticletitle{Object scene flow for autonomous vehicles}. In
  \bibinfo{booktitle}{\emph{IEEE Conference on Computer Vision and Pattern
  Recognition (CVPR)}}. \bibinfo{pages}{3061--3070}.
\newblock


\bibitem[Meuleman et~al\mbox{.}(2023)]%
        {meuleman2023progressively}
\bibfield{author}{\bibinfo{person}{Andreas Meuleman}, \bibinfo{person}{Yu-Lun
  Liu}, \bibinfo{person}{Chen Gao}, \bibinfo{person}{Jia-Bin Huang},
  \bibinfo{person}{Changil Kim}, \bibinfo{person}{Min~H Kim}, {and}
  \bibinfo{person}{Johannes Kopf}.} \bibinfo{year}{2023}\natexlab{}.
\newblock \showarticletitle{Progressively Optimized Local Radiance Fields for
  Robust View Synthesis}. In \bibinfo{booktitle}{\emph{Proc. IEEE Conf. Comput.
  Vis. Pattern Recognit. (CVPR)}}.
\newblock


\bibitem[Mildenhall et~al\mbox{.}(2020)]%
        {mildenhall2020nerf}
\bibfield{author}{\bibinfo{person}{Ben Mildenhall}, \bibinfo{person}{Pratul~P
  Srinivasan}, \bibinfo{person}{Matthew Tancik}, \bibinfo{person}{Jonathan~T
  Barron}, \bibinfo{person}{Ravi Ramamoorthi}, {and} \bibinfo{person}{Ren Ng}.}
  \bibinfo{year}{2020}\natexlab{}.
\newblock \showarticletitle{NeRF: Representing Scenes as Neural Radiance Fields
  for View Synthesis}. In \bibinfo{booktitle}{\emph{Proc. Eur. Conf. Comput.
  Vis. (ECCV)}}.
\newblock


\bibitem[Mildenhall et~al\mbox{.}(2021)]%
        {mildenhall2021nerf}
\bibfield{author}{\bibinfo{person}{Ben Mildenhall}, \bibinfo{person}{Pratul~P
  Srinivasan}, \bibinfo{person}{Matthew Tancik}, \bibinfo{person}{Jonathan~T
  Barron}, \bibinfo{person}{Ravi Ramamoorthi}, {and} \bibinfo{person}{Ren Ng}.}
  \bibinfo{year}{2021}\natexlab{}.
\newblock \showarticletitle{Nerf: Representing scenes as neural radiance fields
  for view synthesis}.
\newblock \bibinfo{journal}{\emph{Commun. ACM}} (\bibinfo{year}{2021}).
\newblock


\bibitem[M{\"u}ller et~al\mbox{.}(2022)]%
        {muller2022instant}
\bibfield{author}{\bibinfo{person}{Thomas M{\"u}ller}, \bibinfo{person}{Alex
  Evans}, \bibinfo{person}{Christoph Schied}, {and} \bibinfo{person}{Alexander
  Keller}.} \bibinfo{year}{2022}\natexlab{}.
\newblock \showarticletitle{Instant Neural Graphics Primitives with a
  Multiresolution Hash Encoding}.
\newblock \bibinfo{journal}{\emph{ACM Trans.~on Graphics (TOG)}}
  (\bibinfo{year}{2022}).
\newblock


\bibitem[Nie et~al\mbox{.}(2019)]%
        {emcua}
\bibfield{author}{\bibinfo{person}{Guang-Yu Nie}, \bibinfo{person}{Ming-Ming
  Cheng}, \bibinfo{person}{Yun Liu}, \bibinfo{person}{Zhengfa Liang},
  \bibinfo{person}{Deng-Ping Fan}, \bibinfo{person}{Yue Liu}, {and}
  \bibinfo{person}{Yongtian Wang}.} \bibinfo{year}{2019}\natexlab{}.
\newblock \showarticletitle{Multi-level context ultra-aggregation for stereo
  matching}. In \bibinfo{booktitle}{\emph{IEEE conference on computer vision
  and pattern recognition (CVPR)}}. \bibinfo{pages}{3283--3291}.
\newblock


\bibitem[Park et~al\mbox{.}(2020)]%
        {park2020non}
\bibfield{author}{\bibinfo{person}{Jinsun Park}, \bibinfo{person}{Kyungdon
  Joo}, \bibinfo{person}{Zhe Hu}, \bibinfo{person}{Chi-Kuei Liu}, {and}
  \bibinfo{person}{In So~Kweon}.} \bibinfo{year}{2020}\natexlab{}.
\newblock \showarticletitle{Non-local spatial propagation network for depth
  completion}. In \bibinfo{booktitle}{\emph{Proc. Eur. Conf. Comput. Vis.
  (ECCV)}}. \bibinfo{pages}{120--136}.
\newblock


\bibitem[Park et~al\mbox{.}(2021)]%
        {park2021nerfies}
\bibfield{author}{\bibinfo{person}{Keunhong Park}, \bibinfo{person}{Utkarsh
  Sinha}, \bibinfo{person}{Jonathan~T Barron}, \bibinfo{person}{Sofien
  Bouaziz}, \bibinfo{person}{Dan~B Goldman}, \bibinfo{person}{Steven~M Seitz},
  {and} \bibinfo{person}{Ricardo Martin-Brualla}.}
  \bibinfo{year}{2021}\natexlab{}.
\newblock \showarticletitle{Nerfies: Deformable neural radiance fields}. In
  \bibinfo{booktitle}{\emph{Proceedings of the IEEE/CVF International
  Conference on Computer Vision}}. \bibinfo{pages}{5865--5874}.
\newblock


\bibitem[Qu et~al\mbox{.}(2020)]%
        {qu2020depth}
\bibfield{author}{\bibinfo{person}{Chao Qu}, \bibinfo{person}{Ty Nguyen}, {and}
  \bibinfo{person}{Camillo Taylor}.} \bibinfo{year}{2020}\natexlab{}.
\newblock \showarticletitle{Depth completion via deep basis fitting}. In
  \bibinfo{booktitle}{\emph{Proc.~of the IEEE Winter Conf.~on Applications of
  Computer Vision (WACV)}}. \bibinfo{pages}{71--80}.
\newblock


\bibitem[Rematas et~al\mbox{.}(2022)]%
        {rematas2022urban}
\bibfield{author}{\bibinfo{person}{Konstantinos Rematas},
  \bibinfo{person}{Andrew Liu}, \bibinfo{person}{Pratul~P Srinivasan},
  \bibinfo{person}{Jonathan~T Barron}, \bibinfo{person}{Andrea Tagliasacchi},
  \bibinfo{person}{Thomas Funkhouser}, {and} \bibinfo{person}{Vittorio
  Ferrari}.} \bibinfo{year}{2022}\natexlab{}.
\newblock \showarticletitle{Urban radiance fields}. In
  \bibinfo{booktitle}{\emph{Proc. IEEE Conf. Comput. Vis. Pattern Recognit.
  (CVPR)}}.
\newblock


\bibitem[Roessle et~al\mbox{.}(2022)]%
        {roessle2022dense}
\bibfield{author}{\bibinfo{person}{Barbara Roessle},
  \bibinfo{person}{Jonathan~T Barron}, \bibinfo{person}{Ben Mildenhall},
  \bibinfo{person}{Pratul~P Srinivasan}, {and} \bibinfo{person}{Matthias
  Nie{\ss}ner}.} \bibinfo{year}{2022}\natexlab{}.
\newblock \showarticletitle{Dense depth priors for neural radiance fields from
  sparse input views}. In \bibinfo{booktitle}{\emph{Proceedings of the IEEE/CVF
  Conference on Computer Vision and Pattern Recognition}}.
  \bibinfo{pages}{12892--12901}.
\newblock


\bibitem[Saxena et~al\mbox{.}(2008)]%
        {mrf}
\bibfield{author}{\bibinfo{person}{Ashutosh Saxena}, \bibinfo{person}{Min Sun},
  {and} \bibinfo{person}{Andrew~Y Ng}.} \bibinfo{year}{2008}\natexlab{}.
\newblock \showarticletitle{Make3d: Learning 3d scene structure from a single
  still image}.
\newblock \bibinfo{journal}{\emph{IEEE Trans.~on Pattern Analalysis and Machine
  Intelligence (TPAMI)}} \bibinfo{volume}{31}, \bibinfo{number}{5}
  (\bibinfo{year}{2008}), \bibinfo{pages}{824--840}.
\newblock


\bibitem[Scharstein and Szeliski(2002)]%
        {scharstein2002taxonomy}
\bibfield{author}{\bibinfo{person}{Daniel Scharstein} {and}
  \bibinfo{person}{Richard Szeliski}.} \bibinfo{year}{2002}\natexlab{}.
\newblock \showarticletitle{A taxonomy and evaluation of dense two-frame stereo
  correspondence algorithms}.
\newblock \bibinfo{journal}{\emph{International journal of computer vision
  (IJCV)}} \bibinfo{volume}{47}, \bibinfo{number}{1-3} (\bibinfo{year}{2002}),
  \bibinfo{pages}{7--42}.
\newblock


\bibitem[Shen et~al\mbox{.}(2021b)]%
        {shen2021learning}
\bibfield{author}{\bibinfo{person}{Guibao Shen}, \bibinfo{person}{Yingkui
  Zhang}, \bibinfo{person}{Jialu Li}, \bibinfo{person}{Mingqiang Wei},
  \bibinfo{person}{Qiong Wang}, \bibinfo{person}{Guangyong Chen}, {and}
  \bibinfo{person}{Pheng-Ann Heng}.} \bibinfo{year}{2021}\natexlab{b}.
\newblock \showarticletitle{Learning regularizer for monocular depth estimation
  with adversarial guidance}. In \bibinfo{booktitle}{\emph{Proceedings of the
  29th ACM International Conference on Multimedia}}.
  \bibinfo{pages}{5222--5230}.
\newblock


\bibitem[Shen et~al\mbox{.}(2021a)]%
        {cfnet}
\bibfield{author}{\bibinfo{person}{Zhelun Shen}, \bibinfo{person}{Yuchao Dai},
  {and} \bibinfo{person}{Zhibo Rao}.} \bibinfo{year}{2021}\natexlab{a}.
\newblock \showarticletitle{Cfnet: Cascade and fused cost volume for robust
  stereo matching}. In \bibinfo{booktitle}{\emph{Proceedings of the IEEE/CVF
  Conference on Computer Vision and Pattern Recognition}}.
  \bibinfo{pages}{13906--13915}.
\newblock


\bibitem[Shen et~al\mbox{.}(2022)]%
        {pcwnet}
\bibfield{author}{\bibinfo{person}{Zhelun Shen}, \bibinfo{person}{Yuchao Dai},
  \bibinfo{person}{Xibin Song}, \bibinfo{person}{Zhibo Rao},
  \bibinfo{person}{Dingfu Zhou}, {and} \bibinfo{person}{Liangjun Zhang}.}
  \bibinfo{year}{2022}\natexlab{}.
\newblock \showarticletitle{Pcw-net: Pyramid combination and warping cost
  volume for stereo matching}. In \bibinfo{booktitle}{\emph{Computer
  Vision--ECCV 2022: 17th European Conference, Tel Aviv, Israel, October
  23--27, 2022, Proceedings, Part XXXII}}. Springer, \bibinfo{pages}{280--297}.
\newblock


\bibitem[Song et~al\mbox{.}(2021)]%
        {lapdepth}
\bibfield{author}{\bibinfo{person}{Minsoo Song}, \bibinfo{person}{Seokjae Lim},
  {and} \bibinfo{person}{Wonjun Kim}.} \bibinfo{year}{2021}\natexlab{}.
\newblock \showarticletitle{Monocular depth estimation using laplacian
  pyramid-based depth residuals}.
\newblock \bibinfo{journal}{\emph{IEEE transactions on circuits and systems for
  video technology}} \bibinfo{volume}{31}, \bibinfo{number}{11}
  (\bibinfo{year}{2021}), \bibinfo{pages}{4381--4393}.
\newblock


\bibitem[Tancik et~al\mbox{.}(2022)]%
        {tancik2022block}
\bibfield{author}{\bibinfo{person}{Matthew Tancik}, \bibinfo{person}{Vincent
  Casser}, \bibinfo{person}{Xinchen Yan}, \bibinfo{person}{Sabeek Pradhan},
  \bibinfo{person}{Ben Mildenhall}, \bibinfo{person}{Pratul~P Srinivasan},
  \bibinfo{person}{Jonathan~T Barron}, {and} \bibinfo{person}{Henrik
  Kretzschmar}.} \bibinfo{year}{2022}\natexlab{}.
\newblock \showarticletitle{Block-nerf: Scalable large scene neural view
  synthesis}. In \bibinfo{booktitle}{\emph{Proc. IEEE Conf. Comput. Vis.
  Pattern Recognit. (CVPR)}}.
\newblock


\bibitem[Tang et~al\mbox{.}(2020)]%
        {tang2020learning}
\bibfield{author}{\bibinfo{person}{Jie Tang}, \bibinfo{person}{Fei-Peng Tian},
  \bibinfo{person}{Wei Feng}, \bibinfo{person}{Jian Li}, {and}
  \bibinfo{person}{Ping Tan}.} \bibinfo{year}{2020}\natexlab{}.
\newblock \showarticletitle{Learning guided convolutional network for depth
  completion}.
\newblock \bibinfo{journal}{\emph{IEEE Trans.~on Image Processing (TIP)}}
  \bibinfo{volume}{30} (\bibinfo{year}{2020}), \bibinfo{pages}{1116--1129}.
\newblock


\bibitem[Tulyakov et~al\mbox{.}(2018)]%
        {pds}
\bibfield{author}{\bibinfo{person}{Stepan Tulyakov}, \bibinfo{person}{Anton
  Ivanov}, {and} \bibinfo{person}{Francois Fleuret}.}
  \bibinfo{year}{2018}\natexlab{}.
\newblock \showarticletitle{Practical deep stereo (pds): Toward
  applications-friendly deep stereo matching}.
\newblock \bibinfo{journal}{\emph{Advances in neural information processing
  systems}}  \bibinfo{volume}{31} (\bibinfo{year}{2018}).
\newblock


\bibitem[Uhrig et~al\mbox{.}(2017)]%
        {uhrig2017sparsity}
\bibfield{author}{\bibinfo{person}{Jonas Uhrig}, \bibinfo{person}{Nick
  Schneider}, \bibinfo{person}{Lukas Schneider}, \bibinfo{person}{Uwe Franke},
  \bibinfo{person}{Thomas Brox}, {and} \bibinfo{person}{Andreas Geiger}.}
  \bibinfo{year}{2017}\natexlab{}.
\newblock \showarticletitle{Sparsity invariant cnns}. In
  \bibinfo{booktitle}{\emph{Proc.~of the Intl.~Conf.~on 3D Vision (3DV)}}.
  \bibinfo{pages}{11--20}.
\newblock


\bibitem[Wang et~al\mbox{.}(2023b)]%
        {wang2023benchmarking}
\bibfield{author}{\bibinfo{person}{Chen Wang}, \bibinfo{person}{Angtian Wang},
  \bibinfo{person}{Junbo Li}, \bibinfo{person}{Alan Yuille}, {and}
  \bibinfo{person}{Cihang Xie}.} \bibinfo{year}{2023}\natexlab{b}.
\newblock \showarticletitle{Benchmarking robustness in neural radiance fields}.
\newblock \bibinfo{journal}{\emph{arXiv preprint arXiv:2301.04075}}
  (\bibinfo{year}{2023}).
\newblock


\bibitem[Wang et~al\mbox{.}(2022)]%
        {wang2022nerf}
\bibfield{author}{\bibinfo{person}{Chen Wang}, \bibinfo{person}{Xian Wu},
  \bibinfo{person}{Yuan-Chen Guo}, \bibinfo{person}{Song-Hai Zhang},
  \bibinfo{person}{Yu-Wing Tai}, {and} \bibinfo{person}{Shi-Min Hu}.}
  \bibinfo{year}{2022}\natexlab{}.
\newblock \showarticletitle{NeRF-SR: High Quality Neural Radiance Fields using
  Supersampling}. In \bibinfo{booktitle}{\emph{Proc.~of the ACM Intl. Conf.~on
  Multimedia. (MM)}}.
\newblock


\bibitem[Wang et~al\mbox{.}(2021)]%
        {wang2021neus}
\bibfield{author}{\bibinfo{person}{Peng Wang}, \bibinfo{person}{Lingjie Liu},
  \bibinfo{person}{Yuan Liu}, \bibinfo{person}{Christian Theobalt},
  \bibinfo{person}{Taku Komura}, {and} \bibinfo{person}{Wenping Wang}.}
  \bibinfo{year}{2021}\natexlab{}.
\newblock \showarticletitle{NeuS: Learning Neural Implicit Surfaces by Volume
  Rendering for Multi-view Reconstruction}. In
  \bibinfo{booktitle}{\emph{Proc.~of the Conference on Neural Information
  Processing Systems (NeurIPS)}}.
\newblock


\bibitem[Wang et~al\mbox{.}(2004)]%
        {wang2004image}
\bibfield{author}{\bibinfo{person}{Zhou Wang}, \bibinfo{person}{Alan~C Bovik},
  \bibinfo{person}{Hamid~R Sheikh}, {and} \bibinfo{person}{Eero~P Simoncelli}.}
  \bibinfo{year}{2004}\natexlab{}.
\newblock \showarticletitle{Image quality assessment: from error visibility to
  structural similarity}.
\newblock \bibinfo{journal}{\emph{IEEE transactions on image processing}}
  \bibinfo{volume}{13}, \bibinfo{number}{4} (\bibinfo{year}{2004}),
  \bibinfo{pages}{600--612}.
\newblock


\bibitem[Wang et~al\mbox{.}(2023a)]%
        {wang2023neural}
\bibfield{author}{\bibinfo{person}{Zian Wang}, \bibinfo{person}{Tianchang
  Shen}, \bibinfo{person}{Jun Gao}, \bibinfo{person}{Shengyu Huang},
  \bibinfo{person}{Jacob Munkberg}, \bibinfo{person}{Jon Hasselgren},
  \bibinfo{person}{Zan Gojcic}, \bibinfo{person}{Wenzheng Chen}, {and}
  \bibinfo{person}{Sanja Fidler}.} \bibinfo{year}{2023}\natexlab{a}.
\newblock \showarticletitle{Neural Fields meet Explicit Geometric
  Representation for Inverse Rendering of Urban Scenes}. In
  \bibinfo{booktitle}{\emph{Proc. IEEE Conf. Comput. Vis. Pattern Recognit.
  (CVPR)}}.
\newblock


\bibitem[Wu et~al\mbox{.}(2023)]%
        {wu23iros}
\bibfield{author}{\bibinfo{person}{Chenming Wu}, \bibinfo{person}{Jiadai Sun},
  \bibinfo{person}{Zhelun Shen}, {and} \bibinfo{person}{Liangjun Zhang}.}
  \bibinfo{year}{2023}\natexlab{}.
\newblock \showarticletitle{MapNeRF: Incorporating Map Priors into Neural
  Radiance Fields for Driving View Simulation}. In
  \bibinfo{booktitle}{\emph{IEEE International Conference on Intelligent Robots
  and Systems (IROS)}}.
\newblock


\bibitem[Wu et~al\mbox{.}(2022)]%
        {wu2022dof}
\bibfield{author}{\bibinfo{person}{Zijin Wu}, \bibinfo{person}{Xingyi Li},
  \bibinfo{person}{Juewen Peng}, \bibinfo{person}{Hao Lu},
  \bibinfo{person}{Zhiguo Cao}, {and} \bibinfo{person}{Weicai Zhong}.}
  \bibinfo{year}{2022}\natexlab{}.
\newblock \showarticletitle{DoF-NeRF: Depth-of-Field Meets Neural Radiance
  Fields}. In \bibinfo{booktitle}{\emph{Proc.~of the ACM Intl. Conf.~on
  Multimedia. (MM)}}. \bibinfo{pages}{1718--1729}.
\newblock


\bibitem[Xie et~al\mbox{.}(2023a)]%
        {anonymous2023snerf}
\bibfield{author}{\bibinfo{person}{Ziyang Xie}, \bibinfo{person}{Junge Zhang},
  \bibinfo{person}{Wenye Li}, \bibinfo{person}{Feihu Zhang}, {and}
  \bibinfo{person}{Li Zhang}.} \bibinfo{year}{2023}\natexlab{a}.
\newblock \showarticletitle{{S-NeRF}: Neural Radiance Fields for Street Views}.
  In \bibinfo{booktitle}{\emph{Proc.~of the Int.~Conf.~on Learning
  Representations (ICLR)}}.
\newblock


\bibitem[Xie et~al\mbox{.}(2023b)]%
        {xie2023s}
\bibfield{author}{\bibinfo{person}{Ziyang Xie}, \bibinfo{person}{Junge Zhang},
  \bibinfo{person}{Wenye Li}, \bibinfo{person}{Feihu Zhang}, {and}
  \bibinfo{person}{Li Zhang}.} \bibinfo{year}{2023}\natexlab{b}.
\newblock \showarticletitle{S-NeRF: Neural Radiance Fields for Street Views}.
\newblock \bibinfo{journal}{\emph{arXiv preprint arXiv:2303.00749}}
  (\bibinfo{year}{2023}).
\newblock


\bibitem[Xing and Chen(2022)]%
        {xing2022mvsplenoctree}
\bibfield{author}{\bibinfo{person}{Wenpeng Xing} {and} \bibinfo{person}{Jie
  Chen}.} \bibinfo{year}{2022}\natexlab{}.
\newblock \showarticletitle{MVSPlenOctree: Fast and Generic Reconstruction of
  Radiance Fields in PlenOctree from Multi-view Stereo}. In
  \bibinfo{booktitle}{\emph{Proc.~of the ACM Intl. Conf.~on Multimedia. (MM)}}.
  \bibinfo{pages}{5114--5122}.
\newblock


\bibitem[Xu and Zhang(2020)]%
        {aanet}
\bibfield{author}{\bibinfo{person}{Haofei Xu} {and} \bibinfo{person}{Juyong
  Zhang}.} \bibinfo{year}{2020}\natexlab{}.
\newblock \showarticletitle{AANet: Adaptive Aggregation Network for Efficient
  Stereo Matching}. In \bibinfo{booktitle}{\emph{IEEE Conference on Computer
  Vision and Pattern Recognition (CVPR)}}. \bibinfo{pages}{1959--1968}.
\newblock


\bibitem[Xu et~al\mbox{.}(2019)]%
        {xu2019depth}
\bibfield{author}{\bibinfo{person}{Yan Xu}, \bibinfo{person}{Xinge Zhu},
  \bibinfo{person}{Jianping Shi}, \bibinfo{person}{Guofeng Zhang},
  \bibinfo{person}{Hujun Bao}, {and} \bibinfo{person}{Hongsheng Li}.}
  \bibinfo{year}{2019}\natexlab{}.
\newblock \showarticletitle{Depth completion from sparse lidar data with
  depth-normal constraints}. In \bibinfo{booktitle}{\emph{Proc.~of the IEEE/CVF
  Intl.~Conf.~on Computer Vision (ICCV)}}. \bibinfo{pages}{2811--2820}.
\newblock


\bibitem[Yang et~al\mbox{.}(2019)]%
        {yang2019hierarchical}
\bibfield{author}{\bibinfo{person}{Gengshan Yang}, \bibinfo{person}{Joshua
  Manela}, \bibinfo{person}{Michael Happold}, {and} \bibinfo{person}{Deva
  Ramanan}.} \bibinfo{year}{2019}\natexlab{}.
\newblock \showarticletitle{Hierarchical deep stereo matching on
  high-resolution images}. In \bibinfo{booktitle}{\emph{Proc. IEEE Conf.
  Comput. Vis. Pattern Recognit. (CVPR)}}. \bibinfo{pages}{5515--5524}.
\newblock


\bibitem[Yang et~al\mbox{.}(2023)]%
        {yang2023unisim}
\bibfield{author}{\bibinfo{person}{Ze Yang}, \bibinfo{person}{Yun Chen},
  \bibinfo{person}{Jingkang Wang}, \bibinfo{person}{Sivabalan Manivasagam},
  \bibinfo{person}{Wei-Chiu Ma}, \bibinfo{person}{Anqi~Joyce Yang}, {and}
  \bibinfo{person}{Raquel Urtasun}.} \bibinfo{year}{2023}\natexlab{}.
\newblock \showarticletitle{UniSim: A Neural Closed-Loop Sensor Simulator}. In
  \bibinfo{booktitle}{\emph{Proceedings of the IEEE/CVF Conference on Computer
  Vision and Pattern Recognition}}. \bibinfo{pages}{1389--1399}.
\newblock


\bibitem[Yu et~al\mbox{.}(2021)]%
        {yu2021pixelnerf}
\bibfield{author}{\bibinfo{person}{Alex Yu}, \bibinfo{person}{Vickie Ye},
  \bibinfo{person}{Matthew Tancik}, {and} \bibinfo{person}{Angjoo Kanazawa}.}
  \bibinfo{year}{2021}\natexlab{}.
\newblock \showarticletitle{pixelnerf: Neural radiance fields from one or few
  images}. In \bibinfo{booktitle}{\emph{Proceedings of the IEEE/CVF Conference
  on Computer Vision and Pattern Recognition}}. \bibinfo{pages}{4578--4587}.
\newblock


\bibitem[Yu et~al\mbox{.}(2022b)]%
        {yu2022pvserf}
\bibfield{author}{\bibinfo{person}{Xianggang Yu}, \bibinfo{person}{Jiapeng
  Tang}, \bibinfo{person}{Yipeng Qin}, \bibinfo{person}{Chenghong Li},
  \bibinfo{person}{Xiaoguang Han}, \bibinfo{person}{Linchao Bao}, {and}
  \bibinfo{person}{Shuguang Cui}.} \bibinfo{year}{2022}\natexlab{b}.
\newblock \showarticletitle{PVSeRF: joint pixel-, voxel-and surface-aligned
  radiance field for single-image novel view synthesis}. In
  \bibinfo{booktitle}{\emph{Proc.~of the ACM Intl. Conf.~on Multimedia. (MM)}}.
  \bibinfo{pages}{1572--1583}.
\newblock


\bibitem[Yu et~al\mbox{.}(2022a)]%
        {yu2022monosdf}
\bibfield{author}{\bibinfo{person}{Zehao Yu}, \bibinfo{person}{Songyou Peng},
  \bibinfo{person}{Michael Niemeyer}, \bibinfo{person}{Torsten Sattler}, {and}
  \bibinfo{person}{Andreas Geiger}.} \bibinfo{year}{2022}\natexlab{a}.
\newblock \showarticletitle{Monosdf: Exploring monocular geometric cues for
  neural implicit surface reconstruction}.
\newblock \bibinfo{journal}{\emph{arXiv preprint arXiv:2206.00665}}
  (\bibinfo{year}{2022}).
\newblock


\bibitem[Zbontar et~al\mbox{.}(2016)]%
        {mccnn}
\bibfield{author}{\bibinfo{person}{Jure Zbontar}, \bibinfo{person}{Yann LeCun},
  {et~al\mbox{.}}} \bibinfo{year}{2016}\natexlab{}.
\newblock \showarticletitle{Stereo matching by training a convolutional neural
  network to compare image patches.}
\newblock \bibinfo{journal}{\emph{J. Mach. Learn. Res.}} \bibinfo{volume}{17},
  \bibinfo{number}{1} (\bibinfo{year}{2016}), \bibinfo{pages}{2287--2318}.
\newblock


\bibitem[Zhang et~al\mbox{.}(2019b)]%
        {ganet}
\bibfield{author}{\bibinfo{person}{Feihu Zhang}, \bibinfo{person}{Victor
  Prisacariu}, \bibinfo{person}{Ruigang Yang}, {and} \bibinfo{person}{Philip~HS
  Torr}.} \bibinfo{year}{2019}\natexlab{b}.
\newblock \showarticletitle{Ga-net: Guided aggregation net for end-to-end
  stereo matching}. In \bibinfo{booktitle}{\emph{IEEE Conference on Computer
  Vision and Pattern Recognition (CVPR)}}. \bibinfo{pages}{185--194}.
\newblock


\bibitem[Zhang et~al\mbox{.}(2022)]%
        {zhang2022vmrf}
\bibfield{author}{\bibinfo{person}{Jiahui Zhang}, \bibinfo{person}{Fangneng
  Zhan}, \bibinfo{person}{Rongliang Wu}, \bibinfo{person}{Yingchen Yu},
  \bibinfo{person}{Wenqing Zhang}, \bibinfo{person}{Bai Song},
  \bibinfo{person}{Xiaoqin Zhang}, {and} \bibinfo{person}{Shijian Lu}.}
  \bibinfo{year}{2022}\natexlab{}.
\newblock \showarticletitle{VMRF: View Matching Neural Radiance Fields}. In
  \bibinfo{booktitle}{\emph{Proc.~of the ACM Intl. Conf.~on Multimedia. (MM)}}.
  \bibinfo{pages}{6579--6587}.
\newblock


\bibitem[Zhang et~al\mbox{.}(2020)]%
        {zhang2020nerf++}
\bibfield{author}{\bibinfo{person}{Kai Zhang}, \bibinfo{person}{Gernot
  Riegler}, \bibinfo{person}{Noah Snavely}, {and} \bibinfo{person}{Vladlen
  Koltun}.} \bibinfo{year}{2020}\natexlab{}.
\newblock \showarticletitle{Nerf++: Analyzing and improving neural radiance
  fields}.
\newblock \bibinfo{journal}{\emph{arXiv preprint arXiv:2010.07492}}
  (\bibinfo{year}{2020}).
\newblock


\bibitem[Zhang et~al\mbox{.}(2018)]%
        {zhang2018unreasonable}
\bibfield{author}{\bibinfo{person}{Richard Zhang}, \bibinfo{person}{Phillip
  Isola}, \bibinfo{person}{Alexei~A Efros}, \bibinfo{person}{Eli Shechtman},
  {and} \bibinfo{person}{Oliver Wang}.} \bibinfo{year}{2018}\natexlab{}.
\newblock \showarticletitle{The unreasonable effectiveness of deep features as
  a perceptual metric}. In \bibinfo{booktitle}{\emph{Proc. IEEE Conf. Comput.
  Vis. Pattern Recognit. (CVPR)}}. \bibinfo{pages}{586--595}.
\newblock


\bibitem[Zhang et~al\mbox{.}(2019a)]%
        {acfnet}
\bibfield{author}{\bibinfo{person}{Youmin Zhang}, \bibinfo{person}{Yimin Chen},
  \bibinfo{person}{Xiao Bai}, \bibinfo{person}{Jun Zhou}, \bibinfo{person}{Kun
  Yu}, \bibinfo{person}{Zhiwei Li}, {and} \bibinfo{person}{Kuiyuan Yang}.}
  \bibinfo{year}{2019}\natexlab{a}.
\newblock \showarticletitle{Adaptive Unimodal Cost Volume Filtering for Deep
  Stereo Matching}. In \bibinfo{booktitle}{\emph{arXiv preprint}}.
\newblock


\bibitem[Zhu et~al\mbox{.}(2023)]%
        {zhu2023nicer}
\bibfield{author}{\bibinfo{person}{Zihan Zhu}, \bibinfo{person}{Songyou Peng},
  \bibinfo{person}{Viktor Larsson}, \bibinfo{person}{Zhaopeng Cui},
  \bibinfo{person}{Martin~R Oswald}, \bibinfo{person}{Andreas Geiger}, {and}
  \bibinfo{person}{Marc Pollefeys}.} \bibinfo{year}{2023}\natexlab{}.
\newblock \showarticletitle{NICER-SLAM: Neural Implicit Scene Encoding for RGB
  SLAM}.
\newblock \bibinfo{journal}{\emph{arXiv preprint arXiv:2302.03594}}
  (\bibinfo{year}{2023}).
\newblock


\end{thebibliography}
\end{document}